\documentclass{article}

\usepackage{arxiv}

\usepackage[utf8]{inputenc} % allow utf-8 input
\usepackage[T1]{fontenc}    % use 8-bit T1 fonts
\usepackage{lmodern}        % https://github.com/rstudio/rticles/issues/343
\usepackage{hyperref}       % hyperlinks
\usepackage{url}            % simple URL typesetting
\usepackage{booktabs}       % professional-quality tables
\usepackage{amsfonts}       % blackboard math symbols
\usepackage{nicefrac}       % compact symbols for 1/2, etc.
\usepackage{microtype}      % microtypography
\usepackage{graphicx}

\title{Random Forests for Time-Fixed and Time-Dependent Predictors: The
DynForest R Package}

\author{
    Anthony Devaux
   \\
    The George Institute for Global Health \\
    UNSW Sydney \\
  Sydney, Australia \\
  \texttt{\href{mailto:anthony.devauxbarault@gmail.com}{\nolinkurl{anthony.devauxbarault@gmail.com}}} \\
   \And
    Cécile Proust-Lima
   \\
    Inserm U1219 - Bordeaux Population Health \\
    Université de Bordeaux \\
  Bordeaux, France \\
  \texttt{\href{mailto:cecile.proust-lima@u-bordeaux.fr}{\nolinkurl{cecile.proust-lima@u-bordeaux.fr}}} \\
   \And
    Robin Genuer
   \\
    Inserm U1219 - Bordeaux Population Health \\
    Université de Bordeaux \\
  Bordeaux, France \\
  \texttt{\href{mailto:robin.genuer@u-bordeaux.fr}{\nolinkurl{robin.genuer@u-bordeaux.fr}}} \\
  }

\usepackage{color}
\usepackage{fancyvrb}

\DefineVerbatimEnvironment{Highlighting}{Verbatim}{commandchars=\\\{\}}
\usepackage{framed}
\definecolor{shadecolor}{RGB}{248,248,248}
\newenvironment{Shaded}{\begin{snugshade}}{\end{snugshade}}

\newcommand{\AttributeTok}[1]{\textcolor[rgb]{0.13,0.29,0.53}{#1}}

\newcommand{\ConstantTok}[1]{\textcolor[rgb]{0.56,0.35,0.01}{#1}}

\newcommand{\DecValTok}[1]{\textcolor[rgb]{0.00,0.00,0.81}{#1}}

\newcommand{\FunctionTok}[1]{\textcolor[rgb]{0.13,0.29,0.53}{\textbf{#1}}}

\newcommand{\NormalTok}[1]{#1}

\newcommand{\OtherTok}[1]{\textcolor[rgb]{0.56,0.35,0.01}{#1}}

\newcommand{\SpecialCharTok}[1]{\textcolor[rgb]{0.81,0.36,0.00}{\textbf{#1}}}

\newcommand{\StringTok}[1]{\textcolor[rgb]{0.31,0.60,0.02}{#1}}

\providecommand{\tightlist}{%
  \setlength{\itemsep}{0pt}\setlength{\parskip}{0pt}}

\usepackage{longtable,booktabs,array}
\usepackage{calc} % for calculating minipage widths
\usepackage{etoolbox}
\makeatletter
\patchcmd\longtable{\par}{\if@noskipsec\mbox{}\fi\par}{}{}
\makeatother
\IfFileExists{footnotehyper.sty}{\usepackage{footnotehyper}}{\usepackage{footnote}}
\makesavenoteenv{longtable}

\newlength{\cslhangindent}
\newlength{\csllabelwidth}
\newlength{\cslentryspacingunit} % times entry-spacing
  {}%
  {\par}
\newenvironment{CSLReferences}[2] % #1 hanging-ident, #2 entry spacing
 {% don't indent paragraphs
  \setlength{\parindent}{0pt}
  \ifodd #1
  \let\oldpar\par
  \def\par{\hangindent=\cslhangindent\oldpar}
  \fi
  \setlength{\parskip}{#2\cslentryspacingunit}
 }%
 {}
\usepackage{calc}

\begin{document}
\maketitle

\begin{abstract}
The R package DynForest implements random forests for predicting a
continuous, a categorical or a (multiple causes) time-to-event outcome
based on time-fixed and time-dependent predictors. The main originality
of DynForest is that it handles time-dependent predictors that can be
endogeneous (i.e., impacted by the outcome process), measured with error
and measured at subject-specific times. At each recursive step of the
tree building process, the time-dependent predictors are internally
summarized into individual features on which the split can be done. This
is achieved using flexible linear mixed models (thanks to the R package
lcmm) which specification is pre-specified by the user. DynForest
returns the mean for continuous outcome, the category with a majority
vote for categorical outcome or the cumulative incidence function over
time for survival outcome. DynForest also computes variable importance
and minimal depth to inform on the most predictive variables or groups
of variables. This paper aims to guide the user with step-by-step
examples for fitting random forests using DynForest.
\end{abstract}

\keywords{
    random forest
   \and
    longitudinal data
   \and
    survival data
   \and
    prediction
  }

\hypertarget{introduction}{%
\section{Introduction}\label{introduction}}

Random forests are a non-parametric powerful method for prediction
purpose. Introduced by Breiman (Breiman 2001) for classification
(categorical outcome) and regression (continuous outcome) frameworks,
random forests are particularly designed to tackle modeling issues in
high-dimension context (\(n << p\)). They can also easily take into
account complex association between the outcome and the predictors
without any pre-specification where regression models are rapidly
limited.

Recently, this methodology was extended to survival data (Hemant
Ishwaran et al. 2008) and competing events (Hemant Ishwaran et al.
2014). Random forests were implemented in several R packages such as
\textbf{randomForestSRC} (H. Ishwaran and Kogalur 2022), \textbf{ranger}
(Wright and Ziegler 2017) or \textbf{xgboost} (Chen and Guestrin 2016)
among others. However, these packages are all limited to time-fixed
predictors. Yet, in many applications, it may be relevant to include
predictors that are repeatedly measured at multiple occasions (regular
or irregular times) with measurement errors to more accurately predict
the outcome. This is the case in particular in health research where an
health outcome is to be predicted according to the history of individual
information.

We developed an original random forests methodology to tackle this issue
and incorporate longitudinal predictors that may be prone-to-error and
possibly intermittently measured (Devaux et al. 2023). The present paper
aims to describe the \textbf{DynForest} R package associated to this
methodology, allowing to predict a continuous, categorical or survival
outcome using multivariate time-dependent predictors.

In section 2, we briefly present \textbf{DynForest} methodology through
its algorithm. In section 3, we present the different functions of
\textbf{DynForest} and we illustrate them in section 4 for a survival
outcome, in section 5 for a categorical outcome and in section 6 for a
continuous outcome. To conclude, we discuss in section 7 the limitations
and future improvements.

\hypertarget{dynforest-principle}{%
\section{DynForest principle}\label{dynforest-principle}}

\textbf{DynForest} is a random forest methodology which can include both
time-fixed predictors of any nature and time-dependent predictors
possibly measured at irregular times. The purpose of \textbf{DynForest}
is to predict an outcome which can be categorical, continuous or
survival (with possibly competing events).

The random forest should first be built on a learning dataset of \(N\)
subjects including: \(Y\) the outcome; \(\mathcal{M}_x\) an ensemble of
\(P\) time-fixed predictors; \(\mathcal{M}_y\) an ensemble of \(Q\)
time-dependent predictors. The random forest consists of an ensemble of
\(B\) trees which are grown as detailed below.

\begin{figure}

{\centering \includegraphics[width=0.8\linewidth,height=0.3\textheight]{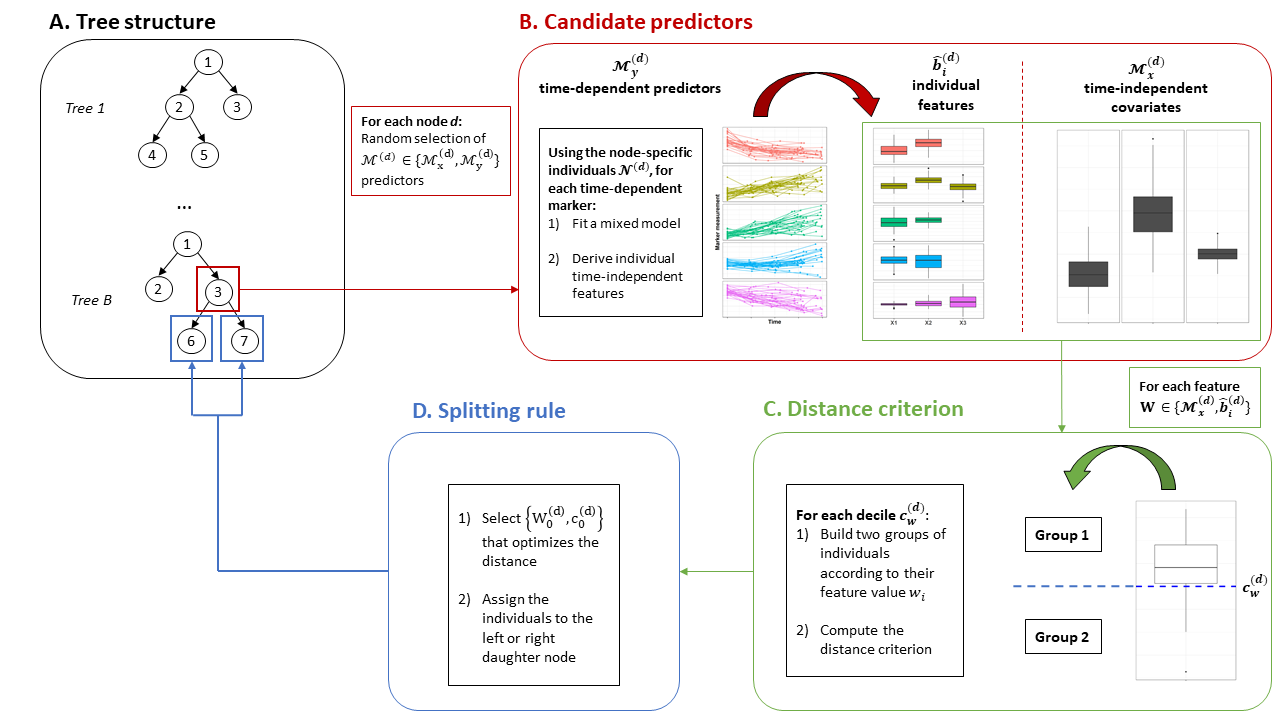} 

}

\caption{Overall scheme of the tree building in DynForest with (A) the tree structure, (B) the node-specific treatment of time-dependent predictors to obtain time-fixed features, (C) the dichotomization of the time-fixed features, (D) the splitting rule.}\label{fig:dynforestRgraph}
\end{figure}

\hypertarget{the-tree-building}{%
\subsection{The tree building}\label{the-tree-building}}

The tree building process, summarized in figure
\ref{fig:dynforestRgraph}, aims to recursively partition the subjects
into groups/nodes that are the most homogeneous regarding the outcome
\(Y\).

For each tree \(b\) \((b = 1, ..., B)\), we first draw a bootstrap
sample from the original dataset of \(N\) subjects (\(N\) draws among
the \(N\) subjects with replacement). The subjects excluded by the
bootstrap constitute the out-of-bag (OOB) sample, noted \(OOB^b\) for
tree \(b\). At each node \(d \in \mathcal{D}^b\) of the tree, we
recursively repeat the following steps using the \(N^{(d)}\) subjects
located at node \(d\):

\begin{enumerate}
\def\labelenumi{\arabic{enumi}.}
\item
  An ensemble of
  \(\mathcal{M}^{(d)}=\{\mathcal{M}_x^{(d)},\mathcal{M}_y^{(d)}\}\)
  candidate predictors are randomly selected among
  \(\{\mathcal{M}_x,\mathcal{M}_y\}\) (see figure
  \ref{fig:dynforestRgraph}B). The size of \(\mathcal{M}^{(d)}\) is
  defined by the hyperparameter \(mtry\).
\item
  For each time-dependent predictor in \(\mathcal{M}_y^{(d)}\):

  \begin{enumerate}
  \def\labelenumii{\alph{enumii}.}
  \tightlist
  \item
    We independently model the trajectory of the predictor using a
    flexible linear mixed model (Laird and Ware 1982) according to time
    (the specification of the model is defined by the user)
  \item
    We derive an ensemble \(\mathcal{M}_{y\star}^{(d)}\) of individual
    time-independent features. These features are the individual
    random-effects of the linear mixed model predicted from the repeated
    data of individual \(i = 1, ..., N^{(d)}\)
  \end{enumerate}
\item
  We define
  \(\mathcal{M}_\star^{(d)}=\{\mathcal{M}_x^{(d)},\mathcal{M}_{y \star}^{(d)}\}\)
  our new ensemble of candidate features.
\item
  For each candidate feature \(W \in \mathcal{M}_\star^{(d)}\):

  \begin{enumerate}
  \def\labelenumii{\alph{enumii}.}
  \tightlist
  \item
    We build a serie of splits \(c_W^{(d)}\) according to the feature
    values if continuous, or subsets of categories otherwise (see figure
    \ref{fig:dynforestRgraph}C), leading each time to two groups.
  \item
    We quantify the distance between the two groups according to the
    nature of \(Y\):

    \begin{itemize}
    \tightlist
    \item
      If \(Y\) continuous: we compute the weighted within-group variance
      with the proportion of subjects in each group as weights
    \item
      If \(Y\) categorical: we compute the weighted within-group Shannon
      entropy (Shannon 1948) (i.e., the amount of uncertainty) with the
      proportion of subjects in each group as weights
    \item
      If \(Y\) survival \underline{without} competing events: we compute
      the log-rank statistic test (Peto and Peto 1972)
    \item
      If \(Y\) survival \underline{with} competing events: we compute
      the Fine \& Gray statistic test (R. J. Gray 1988)
    \end{itemize}
  \end{enumerate}
\item
  We split the subjects into the two groups that minimize (for
  continuous and categorical outcome) or maximize (for survival outcome)
  the quantity defined previously. We denote \(\{W_0^d,c_0^d\}\) the
  optimal couple used to split the subjects and assign them to the left
  and right daughter nodes \(2d\) and \(2d + 1\), respectively (see
  figure \ref{fig:dynforestRgraph}D and A).
\item
  Step 1 to 5 are iterated on the daughter nodes until stopping criteria
  are met.
\end{enumerate}

We define two stopping criteria: \texttt{nodesize} the minimal number of
subjects in a node required to reiterate the split and \texttt{minsplit}
the minimal number of events required to split the node.
\texttt{minsplit} is only defined with survival outcome. In the
following, we call leaves the terminal nodes.

In each leaf \(h \in \mathcal{H}^b\) of tree \(b\), a summary
\(\pi^{h^b}\) is computed using the individuals belonging to the leaf.
The leaf summary is defined according to the outcome:

\begin{itemize}
\tightlist
\item
  the mean, for \(Y\) continuous
\item
  the category with the highest probability, for \(Y\) categorical
\item
  the cumulative incidence function over time computed using the
  Nelson-Aalen cumulative hazard function estimator (Nelson 1969; O.
  Aalen 1976), for \(Y\) single cause time-to-event
\item
  the cumulative incidence function over time computed using the
  non-parametric Aalen-Johansen estimator (O. O. Aalen and Johansen
  1978), for \(Y\) time-to-event with multiple causes
\end{itemize}

\hypertarget{individual-prediction-of-the-outcome}{%
\subsection{Individual prediction of the
outcome}\label{individual-prediction-of-the-outcome}}

\hypertarget{out-of-bag-individual-prediction}{%
\subsubsection{Out-Of-Bag individual
prediction}\label{out-of-bag-individual-prediction}}

The overall OOB prediction \(\hat{\pi}_{ \star}\) for a subject
\(\star\) can be computed by averaging the tree-based predictions of
\(\star\) over the random forest as follows: \begin{equation}
  \hat{\pi}_{ \star} = \frac{1}{|\mathcal{O}_\star|} \sum_{b \in \mathcal{O}_\star} \hat{\pi}^{h_\star^b}
\end{equation} where \(\mathcal{O}_\star\) is the ensemble of trees
where \(\star\) is \(OOB\) and \(|\mathcal{O}_\star|\) denotes its
cardinal The prediction \(\hat{\pi}^{h_\star^b}\) is obtained by
dropping down subject \(\star\) along tree \(b\). At each node
\(d \in \mathcal{D}^b\), the subject \(\star\) is assigned to the left
or right node according to his/her data and the optimal couple
\(\{W_0^d,c_0^d\}\). \(W_0^d\) is a random-effect feature, its value for
\(\star\) is predicted from the individual repeated measures using the
estimated parameters from the linear mixed model.

\hypertarget{individual-dynamic-prediction-from-a-landmark-time}{%
\subsubsection{Individual dynamic prediction from a landmark
time}\label{individual-dynamic-prediction-from-a-landmark-time}}

With a survival outcome, the OOB prediction described in the previous
paragraph can be extended to compute the individual probability of event
from a landmark time \(s\) by exploiting the repeated measures of
subject \(\star\) only until \(s\). For a new subject \(\star\), we thus
define the individual prediction \(\hat{\pi}_{\star}(s)\) at landmark
time \(s\) with: \begin{equation}
  \hat{\pi}_{\star}(s) = \frac{1}{B} \sum_{b=1}^B \hat{\pi}^{h_\star^b}(s)
\end{equation} where \(\hat{\pi}^{h_\star^b}(s)\) is the tree-based
prediction computed by dropping down \(\star\) along the tree by
considering longitudinal predictors collected until \(s\) and time-fixed
predictors.

\hypertarget{out-of-bag-prediction-error}{%
\subsection{Out-Of-Bag prediction
error}\label{out-of-bag-prediction-error}}

Using the OOB individual predictions, an OOB prediction error can be
internally assessed. The OOB prediction error quantifies the difference
between the observed and the predicted values. It is defined according
to the nature of \(Y\) as:

\begin{itemize}
\tightlist
\item
  for \(Y\) continuous, the mean square error (MSE) defined by:
\end{itemize}

\begin{equation}
  errOOB = \frac{1}{N} \sum_{i=1}^N ( \hat{\pi}_i - \pi_i^0 )^2
\end{equation}

\begin{itemize}
\tightlist
\item
  for \(Y\) categorical, the missclassification error defined by:
\end{itemize}

\begin{equation}
  errOOB = \frac{1}{N} \sum_{i=1}^N {1}_{( \hat{\pi}_i \neq \pi_i^0 )}
\end{equation}

\begin{itemize}
\tightlist
\item
  for \(Y\) survival, the Integrated Brier Score (IBS) (Sène et al.
  2016) between \(\tau_1\) and \(\tau_2\) defined by:
\end{itemize}

\begin{equation}
  errOOB = \int_{\tau_1}^{\tau_2} \frac{1}{N} \sum_{i=1}^{N}  \hat{\omega}_i(t) \Big\{ I(T_i \leq t, \delta_i = k) - 
  \hat{\pi}_{ik}(t) \Big) \Big\}^2 dt
\end{equation}

with \(T\) the time-to-event, \(k\) the cause of interest and
\(\hat{\omega}(t)\) the estimated weights using Inverse Probability of
Censoring Weights (IPCW) technique that accounts for censoring (Gerds
and Schumacher 2006).

The OOB error of prediction is used to particular to tune the random
forest by determining the hyperparameters (i.e., \texttt{mtry},
\texttt{nodesize} and \texttt{minsplit}) which give the smallest OOB
prediction error.

\hypertarget{explore-the-most-predictive-variables}{%
\subsection{Explore the most predictive
variables}\label{explore-the-most-predictive-variables}}

\hypertarget{variable-importance}{%
\subsubsection{Variable importance}\label{variable-importance}}

The variable importance (VIMP) measures the loss of predictive
performance (Hemant Ishwaran et al. 2008) when removing the link between
a predictor and the outcome. The link is removed by permuting the
predictor values at the subject level for time-fixed predictors or at
observation level for time-dependent predictors. A large VIMP value
indicates a good prediction ability for the predictor.

However, in case of correlated predictors, the VIMP may not properly
quantify the variable importance (Gregorutti, Michel, and Saint-Pierre
2017) as the information of the predictor may still be present. To
better handle situations with highly correlated predictors, the grouped
variable importance (gVIMP) can be computed indirectly. It consists in
simultaneously evaluate the importance of a group of predictors defined
by the user. The computation is the same as for the VIMP except the
permutation is performed simultaneously on all the predictors of the
group. A large gVIMP value indicates a good prediction ability for the
group of predictors.

\hypertarget{minimal-depth}{%
\subsubsection{Minimal depth}\label{minimal-depth}}

The minimal depth is another statistic to quantify the importance of a
variable. It assesses the distance between the root node and the first
node for which the predictor is used to split the subjects (1 for first
level, 2 for second level, 3 for third level, \ldots). This statistic
can be computed at the predictor level or at the feature level, allowing
to fully understand the tree building process.

We strongly advice to compute the minimal depth with \texttt{mtry}
hyperparameter chosen at its maximum to ensure that all predictors are
systematically among candidate predictors for splitting the subjects.

\hypertarget{the-dynforest-r-package}{%
\section{\texorpdfstring{The \textbf{DynForest} R
package}{The DynForest R package}}\label{the-dynforest-r-package}}

\textbf{DynForest} methodology was implemented into the R package
\textbf{DynForest} (Devaux 2024) freely available on The Comprehensive R
Archive Network (CRAN) to users.

The package includes two main functions: \texttt{DynForest()} and
\texttt{predict()} for the learning and the prediction steps. These
functions are fully described in section 3.1 and 3.2. Other functions
available are briefly described in the table below. These functions are
illustrated in examples, one for a survival outcome, one for a
categorical outcome and one for a continuous outcome.

\begin{longtable}[]{@{}
  >{\raggedright\arraybackslash}p{(\columnwidth - 2\tabcolsep) * \real{0.5000}}
  >{\raggedright\arraybackslash}p{(\columnwidth - 2\tabcolsep) * \real{0.5000}}@{}}
\toprule\noalign{}
\begin{minipage}[b]{\linewidth}\raggedright
Function
\end{minipage} & \begin{minipage}[b]{\linewidth}\raggedright
Description
\end{minipage} \\
\midrule\noalign{}
\endhead
\bottomrule\noalign{}
\endlastfoot
\emph{Learning and prediction steps} & \\
\texttt{DynForest()} & Function that builds the random forest \\
\texttt{predict()} & Function for S3 class \texttt{DynForest} predicting
the outcome on new subjects using the individual-specific information \\
\emph{Assessment function} & \\
\texttt{compute\_OOBerror()} & Function that computes the Out-Of-Bag
error to be minimized to tune the random forest \\
\emph{Exploring functions} & \\
\texttt{compute\_VIMP()} & Function that computes the importance of
variables \\
\texttt{compute\_gVIMP()} & Function that computes the importance of a
group of variables \\
\texttt{var\_depth()} & Function that extracts information about the
tree building process \\
\emph{plot() functions for S3 class:} & \\
\texttt{DynForest} & Plot the estimated CIF for given tree nodes or
subjects \\
\texttt{DynForestPred} & Plot the predicted CIF for the cause of
interest for given subjects \\
\texttt{DynForestVIMP} & Plot the importance of variables by value or
percentage \\
\texttt{DynForestgVIMP} & Plot the importance of a group of variables by
value or percentage \\
\texttt{DynForestVarDepth} & Plot the minimal depth by predictors or
features \\
\emph{Other functions} & \\
\texttt{summary()} & Function for class S3 \texttt{DynForest} or
\texttt{DynForestOOB} displaying information about the type of random
forest, predictors included, parameters used, Out-Of-Bag error (only for
\texttt{DynForestOOB} class) and brief summaries about the leaves \\
\texttt{print()} & Function to print object of class \texttt{DynForest},
\texttt{DynForestOOB}, \texttt{DynForestVIMP}, \texttt{DynForestgVIMP},
\texttt{DynForestVarDepth} and \texttt{DynForestPred} \\
\texttt{getTree()} & Function that extracts the tree structure for a
given tree \\
\texttt{getTreeNode()} & Function that extracts the terminal node
identifiers for a given tree \\
\end{longtable}

\hypertarget{dynforest-function}{%
\subsection{\texorpdfstring{\texttt{DynForest()}
function}{DynForest() function}}\label{dynforest-function}}

\texttt{DynForest()} is the function to build the random forest. The
call of this function is:

\begin{Shaded}
\begin{Highlighting}[]
\FunctionTok{DynForest}\NormalTok{(}\AttributeTok{timeData =} \ConstantTok{NULL}\NormalTok{, }\AttributeTok{fixedData =} \ConstantTok{NULL}\NormalTok{, }\AttributeTok{idVar =} \ConstantTok{NULL}\NormalTok{, }
          \AttributeTok{timeVar =} \ConstantTok{NULL}\NormalTok{, }\AttributeTok{timeVarModel =} \ConstantTok{NULL}\NormalTok{, }\AttributeTok{Y =} \ConstantTok{NULL}\NormalTok{, }
          \AttributeTok{ntree =} \DecValTok{200}\NormalTok{, }\AttributeTok{mtry =} \ConstantTok{NULL}\NormalTok{, }\AttributeTok{nodesize =} \DecValTok{1}\NormalTok{, }\AttributeTok{minsplit =} \DecValTok{2}\NormalTok{, }\AttributeTok{cause =} \DecValTok{1}\NormalTok{,}
          \AttributeTok{nsplit\_option =} \StringTok{"quantile"}\NormalTok{, }\AttributeTok{ncores =} \ConstantTok{NULL}\NormalTok{,}
          \AttributeTok{seed =} \DecValTok{1234}\NormalTok{, }\AttributeTok{verbose =} \ConstantTok{TRUE}\NormalTok{)}
\end{Highlighting}
\end{Shaded}

\hypertarget{arguments}{%
\subsubsection{Arguments}\label{arguments}}

\texttt{timeData} is an optional argument that contains the dataframe in
longitudinal format (i.e., one observation per row) for the
time-dependent predictors. In addition to time-dependent predictors,
this dataframe should include a unique identifier and the measurement
times. This argument is set to \texttt{NULL} if no time-dependent
predictor is included. Argument \texttt{fixedData} contains the
dataframe in wide format (i.e., one subject per row) for the time-fixed
predictors. In addition to time-fixed predictors, this dataframe should
also include the same identifier as used in \texttt{timeData.} This
argument is set to \texttt{NULL} if no time-fixed predictor is included.
Argument \texttt{idVar} provides the name of identifier variable
included in \texttt{timeData} and \texttt{fixedData} dataframes.
Argument \texttt{timeVar} provides the name of time variable included in
\texttt{timeData} dataframe. Argument \texttt{timeVarModel} contains a
list that specifies the structure of the mixed models assumed for each
longitudinal predictor of \texttt{timeData} dataframe. For each
longitudinal predictor, the list should contain a \texttt{fixed} and a
\texttt{random} argument to define the formula of a mixed model to be
estimated with \textbf{lcmm} R package (Proust-Lima, Philipps, and
Liquet 2017). \texttt{fixed} defines the formula for the fixed-effects
and \texttt{random} for the random-effects. Argument \texttt{Y} contains
a list of two elements \texttt{type} and \texttt{Y}. Element
\texttt{type} defines the nature of the outcome (\texttt{surv} for
survival outcome with possibly competing causes, \texttt{numeric} for
continuous outcome and \texttt{factor} for categorical outcome) and
element \texttt{Y} defines the dataframe which includes the identifier
(same as in \texttt{timeData} and \texttt{fixedData} dataframes) and
outcome variables.

Arguments \texttt{ntree}, \texttt{mtry}, \texttt{nodesize} and
\texttt{minsplit} are the hyperparameters of the random forest. Argument
\texttt{ntree} controls the number of trees in the random forest (200 by
default). Argument \texttt{mtry} indicates the number of variables
randomly drawn at each node (square root of the total number of
predictors by default). Argument \texttt{nodesize} indicates the minimal
number of subjects allowed in the leaves (1 by default). Argument
\texttt{minsplit} controls the minimal number of events required to
split the node (2 by default).

For survival outcome, argument \texttt{cause} indicates the event the
interest. Argument \texttt{nsplit\_option} indicates the method to build
the two groups of individuals at each node. By default, we build the
groups according to deciles (\texttt{quantile} option) but they could be
built according to random values (\texttt{sample} option).

Argument \texttt{ncores} indicates the number of cores used to grow the
trees in parallel mode. By default, we set the number of cores of the
computer minus 1. Argument \texttt{seed} specifies the random seed. It
can be fixed to replicate the results. Argument \texttt{verbose} allows
to display a progression bar during the execution of the function.

\hypertarget{values}{%
\subsubsection{Values}\label{values}}

\texttt{DynForest()} function returns an object of class
\texttt{DynForest} containing several elements:

\begin{itemize}
\tightlist
\item
  \texttt{data} a list with longitudinal predictors
  (\texttt{Longitudinal} element), continuous predictors
  (\texttt{Numeric} element) and categorical predictors (\texttt{Factor}
  element)
\item
  \texttt{rf} is a dataframe with one column per tree containing a list
  with several elements, which includes:

  \begin{itemize}
  \tightlist
  \item
    \texttt{leaves} the leaf identifier for each subject used to grow
    the tree
  \item
    \texttt{idY} the identifiers for each subject used to grow the tree
  \item
    \texttt{V\_split} the split summary (more detailed below)
  \item
    \texttt{Y\_pred} the estimated outcome in each leaf
  \item
    \texttt{model\_param} the estimated parameters of the mixed model
    for the longitudinal predictors used to split the subjects at each
    node
  \item
    \texttt{Ytype}, \texttt{hist\_nodes}, \texttt{Y}, \texttt{boot} and
    \texttt{Ylevels} internal information used in other functions
  \end{itemize}
\item
  \texttt{type} the nature of the outcome
\item
  \texttt{times} the event times (only for survival outcome)
\item
  \texttt{cause} the cause of interest (only for survival outcome)
\item
  \texttt{causes} the unique causes (only for survival outcome)
\item
  \texttt{Inputs} the list of predictors names for \texttt{Longitudinal}
  (longitudinal predictor), \texttt{Continuous} (continuous predictor)
  and \texttt{Factor} (categorical predictor)
\item
  \texttt{Longitudinal.model} the mixed model specification for each
  longitudinal predictor
\item
  \texttt{param} a list of hyperparameters used to grow the random
  forest
\item
  \texttt{comput.time} the computation time
\end{itemize}

\noindent The main information returned by \texttt{rf} is
\texttt{V\_split} element which can also be extract using
\texttt{getTree()} function. This element contains a table sorted by the
node/leaf identifier (\texttt{id\_node} column) with each row
representing a node/leaf. Each column provides information about the
splits:

\begin{itemize}
\tightlist
\item
  \texttt{type}: the nature of the predictor (\texttt{Longitudinal} for
  longitudinal predictor, \texttt{Numeric} for continuous predictor or
  \texttt{Factor} for categorical predictor) if the node was split,
  \texttt{Leaf} otherwise;
\item
  \texttt{var\_split}: the predictor used for the split defined by its
  order in \texttt{timeData} and \texttt{fixedData};
\item
  \texttt{feature}: the feature used for the split defined by its
  position in random statistic;
\item
  \texttt{threshold}: the threshold used for the split (only with
  \texttt{Longitudinal} and \texttt{Numeric}). No information is
  returned for \texttt{Factor};
\item
  \texttt{N}: the number of subjects in the node/leaf;
\item
  \texttt{Nevent}: the number of events of interest in the node/leaf
  (only with survival outcome);
\item
  \texttt{depth}: the depth level of the node/leaf.
\end{itemize}

\hypertarget{additional-information-about-the-dependencies}{%
\subsubsection{Additional information about the
dependencies}\label{additional-information-about-the-dependencies}}

\texttt{DynForest()} function internally calls other functions from
related packages to build the random forest:

\begin{itemize}
\tightlist
\item
  \texttt{hlme()} function (from \textbf{lcmm} package (Proust-Lima,
  Philipps, and Liquet 2017)) to fit the mixed models for the
  time-dependent predictors defined in \texttt{timeData} and
  \texttt{timeVarModel} arguments
\item
  \texttt{Entropy()} function (from \textbf{base} package) to compute
  the Shannon entropy
\item
  \texttt{survdiff()} function (from \textbf{survival} package (Therneau
  2022)) to compute the log-rank statistic test
\item
  \texttt{crr()} function (from \textbf{cmprsk} package (B. Gray 2020))
  to compute the Fine \& Gray statistic test
\end{itemize}

\hypertarget{predict-function}{%
\subsection{\texorpdfstring{\texttt{predict()}
function}{predict() function}}\label{predict-function}}

\texttt{predict()} is the S3 function for class \texttt{DynForest} to
predict the outcome on new subjects. Landmark time can be specified to
consider only longitudinal data collected up to this time to compute the
prediction. The call of this function is:

\begin{Shaded}
\begin{Highlighting}[]
\FunctionTok{predict}\NormalTok{(object, }\AttributeTok{timeData =} \ConstantTok{NULL}\NormalTok{, }\AttributeTok{fixedData =} \ConstantTok{NULL}\NormalTok{,}
\NormalTok{        idVar, timeVar, }\AttributeTok{t0 =} \ConstantTok{NULL}\NormalTok{)}
\end{Highlighting}
\end{Shaded}

\hypertarget{arguments-1}{%
\subsubsection{Arguments}\label{arguments-1}}

Argument \texttt{object} contains a \texttt{DynForest} object resulting
from \texttt{DynForest()} function. Argument \texttt{timeData} contains
the dataframe in longitudinal format (i.e., one observation per row) for
the time-dependent predictors of new subjects. In addition to
time-dependent predictors, this dataframe should also include a unique
identifier and the time measurements. This argument can be set to
\texttt{NULL} if no time-dependent predictor is included. Argument
\texttt{fixedData} contains the dataframe in wide format (i.e., one
subject per row) for the time-fixed predictors of new subjects. In
addition to time-fixed predictors, this dataframe should also include an
unique identifier. This argument can be set to \texttt{NULL} if no
time-fixed predictor is included. Argument \texttt{idVar} provides the
name of the identifier variable included in \texttt{timeData} and
\texttt{fixedData} dataframes. Argument \texttt{timeVar} provides the
name of time-measurement variable included in \texttt{timeData}
dataframe. Argument \texttt{t0} defines the landmark time; only the
longitudinal data collected up to this time are to be considered. This
argument should be set to \texttt{NULL} to include all longitudinal
data.

\hypertarget{values-1}{%
\subsubsection{Values}\label{values-1}}

\texttt{predict()} function returns several elements:

\begin{itemize}
\tightlist
\item
  \texttt{t0} the landmark time defined in argument (\texttt{NULL} by
  default)
\item
  \texttt{times} times used to compute the individual predictions (only
  with survival outcome). The times are defined according to the
  time-to-event subjects used to build the random forest.
\item
  \texttt{pred\_indiv} the predicted outcome for the new subject. With
  survival outcome, predictions are provided for each time defined in
  \texttt{times} element.
\item
  \texttt{pred\_leaf} a table giving for each tree (in column) the leaf
  in which each subject is assigned (in row)
\item
  \texttt{pred\_indiv\_proba} the proportion of the trees leading to the
  category prediction for each subject (only with categorical outcome)
\end{itemize}

\hypertarget{sec:dynforestR_surv}{%
\section{How to use DynForest R package with a survival
outcome?}\label{sec:dynforestR_surv}}

\hypertarget{illustrative-dataset-pbc2-dataset}{%
\subsection{\texorpdfstring{Illustrative dataset: \texttt{pbc2}
dataset}{Illustrative dataset: pbc2 dataset}}\label{illustrative-dataset-pbc2-dataset}}

The \texttt{pbc2} dataset (Murtaugh et al. 1994) is loaded with the
package \textbf{DynForest} to illustrate its function abilities.
\texttt{pbc2} data come from a clinical trial conducted by the Mayo
Clinic between 1974 and 1984 to treat the primary biliary cholangitis
(PBC), a chronic liver disease. 312 patients were enrolled in a clinical
trial to evaluate the effectiveness of D-penicillamine compared to a
placebo to treat the PBC and were followed since the clinical trial
ends, leading to a total of 1945 observations. During the follow-up,
several clinical continuous markers were collected over time such as:
the level of serum bilirubin (\texttt{serBilir}), the level of serum
cholesterol (\texttt{serChol}), the level of albumin (\texttt{albumin}),
the level of alkaline (\texttt{alkaline}), the level of aspartate
aminotransferase (\texttt{SGOT}), platelets count (\texttt{platelets})
and the prothrombin time (\texttt{prothrombin}). 4 non-continuous
time-dependent predictors were also collected: the presence of ascites
(\texttt{ascites}), the presence of hepatomegaly
(\texttt{hepatomegaly}), the presence of blood vessel malformations in
the skin (\texttt{spiders}) and the edema levels (\texttt{edema}). These
time-dependent predictors were recorded according to \texttt{time}
variable. In addition to these time-dependent predictors, few predictors
were collected at enrollment: the sex (\texttt{sex}), the age
(\texttt{age}) and the drug treatment (\texttt{drug}). During the
follow-up, 140 patients died before transplantation, 29 patients were
transplanted and 143 patients were censored alive (\texttt{event}). The
time of first event (censored alive or any event) was considered as the
event time (\texttt{years})

\begin{verbatim}
##    id      time ascites hepatomegaly spiders                   edema serBilir
## 1   1 0.0000000     Yes          Yes     Yes edema despite diuretics     14.5
## 2   1 0.5256817     Yes          Yes     Yes edema despite diuretics     21.3
## 3  10 0.0000000     Yes           No     Yes edema despite diuretics     12.6
## 4 100 0.0000000      No          Yes      No                No edema      2.3
## 5 100 0.4681853      No          Yes      No                No edema      2.5
## 6 100 0.9801774     Yes           No      No      edema no diuretics      2.9
##   serChol albumin alkaline  SGOT platelets prothrombin histologic      drug
## 1     261    2.60     1718 138.0       190        12.2          4 D-penicil
## 2      NA    2.94     1612   6.2       183        11.2          4 D-penicil
## 3     200    2.74      918 147.3       302        11.5          4   placebo
## 4     178    3.00      746 178.3       119        12.0          4   placebo
## 5      NA    2.94      836 189.1        98        11.4          4   placebo
## 6      NA    3.02      650 124.0        99        11.7          4   placebo
##        age    sex     years event
## 1 58.76684 female 1.0951703     2
## 2 58.76684 female 1.0951703     2
## 3 70.56182 female 0.1396342     2
## 4 51.47027   male 1.5113350     2
## 5 51.47027   male 1.5113350     2
## 6 51.47027   male 1.5113350     2
\end{verbatim}

For the illustration, 4 time-dependent predictors (\texttt{serBilir},
\texttt{SGOT}, \texttt{albumin} and \texttt{alkaline}) and 3 predictors
measured at enrollment (\texttt{sex}, \texttt{age} and \texttt{drug})
were considered. We aim to predict the death without transplantation on
patients suffering from primary biliary cholangitis (PBC) using clinical
and socio-demographic predictors, considering the transplantation as a
competing event.

\hypertarget{data-management}{%
\subsection{Data management}\label{data-management}}

To begin, we split the subjects into two datasets: (i) one dataset to
train the random forest using \(2/3\) of patients; (ii) one dataset to
predict on the other \(1/3\) of patients. The random seed is set to 1234
for replication purpose.

\begin{Shaded}
\begin{Highlighting}[]
\FunctionTok{set.seed}\NormalTok{(}\DecValTok{1234}\NormalTok{)}
\NormalTok{id }\OtherTok{\textless{}{-}} \FunctionTok{unique}\NormalTok{(pbc2}\SpecialCharTok{$}\NormalTok{id)}
\NormalTok{id\_sample }\OtherTok{\textless{}{-}} \FunctionTok{sample}\NormalTok{(id, }\FunctionTok{length}\NormalTok{(id)}\SpecialCharTok{*}\DecValTok{2}\SpecialCharTok{/}\DecValTok{3}\NormalTok{)}
\NormalTok{id\_row }\OtherTok{\textless{}{-}} \FunctionTok{which}\NormalTok{(pbc2}\SpecialCharTok{$}\NormalTok{id }\SpecialCharTok{\%in\%}\NormalTok{ id\_sample)}
\NormalTok{pbc2\_train }\OtherTok{\textless{}{-}}\NormalTok{ pbc2[id\_row,]}
\NormalTok{pbc2\_pred }\OtherTok{\textless{}{-}}\NormalTok{ pbc2[}\SpecialCharTok{{-}}\NormalTok{id\_row,]}
\end{Highlighting}
\end{Shaded}

Then, we build the dataframe \texttt{timeData\_train} in the
longitudinal format (i.e., one observation per row) for the longitudinal
predictors including: \texttt{id} the unique patient identifier;
\texttt{time} the observed time measurements; \texttt{serBilir},
\texttt{SGOT}, \texttt{albumin} and \texttt{alkaline} the longitudinal
predictors. We also build the dataframe \texttt{fixedData\_train} with
the time-fixed predictors including: \texttt{id} the unique patient
identifier; \texttt{age}, \texttt{drug} and \texttt{sex} predictors
measured at enrollment. The nature of each predictor needs to be
properly defined with \texttt{as.factor()} function for categorical
predictors (e.g., \texttt{drug} and \texttt{sex}).

\begin{Shaded}
\begin{Highlighting}[]
\NormalTok{timeData\_train }\OtherTok{\textless{}{-}}\NormalTok{ pbc2\_train[,}\FunctionTok{c}\NormalTok{(}\StringTok{"id"}\NormalTok{,}\StringTok{"time"}\NormalTok{,}
                                \StringTok{"serBilir"}\NormalTok{,}\StringTok{"SGOT"}\NormalTok{,}
                                \StringTok{"albumin"}\NormalTok{,}\StringTok{"alkaline"}\NormalTok{)]}
\NormalTok{fixedData\_train }\OtherTok{\textless{}{-}} \FunctionTok{unique}\NormalTok{(pbc2\_train[,}\FunctionTok{c}\NormalTok{(}\StringTok{"id"}\NormalTok{,}\StringTok{"age"}\NormalTok{,}\StringTok{"drug"}\NormalTok{,}\StringTok{"sex"}\NormalTok{)])}
\end{Highlighting}
\end{Shaded}

\hypertarget{specification-of-the-models-for-the-time-dependent-predictors}{%
\subsection{Specification of the models for the time-dependent
predictors}\label{specification-of-the-models-for-the-time-dependent-predictors}}

The first step of the random forest building consists in specify the
mixed model of each longitudinal predictor through a list containing the
fixed and random formula for the fixed effect and random effects of the
mixed models, respectively. Here, we assume a linear trajectory for
\texttt{serBilir}, \texttt{albumin} and \texttt{alkaline}, and quadratic
trajectory for \texttt{SGOT.} Although, we restricted this example to
linear and quadratic functions of time, we note that any function can be
considered including splines.

\begin{Shaded}
\begin{Highlighting}[]
\NormalTok{timeVarModel }\OtherTok{\textless{}{-}} \FunctionTok{list}\NormalTok{(}\AttributeTok{serBilir =} \FunctionTok{list}\NormalTok{(}\AttributeTok{fixed =}\NormalTok{ serBilir }\SpecialCharTok{\textasciitilde{}}\NormalTok{ time,}
                                     \AttributeTok{random =} \SpecialCharTok{\textasciitilde{}}\NormalTok{ time),}
                     \AttributeTok{SGOT =} \FunctionTok{list}\NormalTok{(}\AttributeTok{fixed =}\NormalTok{ SGOT }\SpecialCharTok{\textasciitilde{}}\NormalTok{ time }\SpecialCharTok{+} \FunctionTok{I}\NormalTok{(time}\SpecialCharTok{\^{}}\DecValTok{2}\NormalTok{),}
                                 \AttributeTok{random =} \SpecialCharTok{\textasciitilde{}}\NormalTok{ time }\SpecialCharTok{+} \FunctionTok{I}\NormalTok{(time}\SpecialCharTok{\^{}}\DecValTok{2}\NormalTok{)),}
                     \AttributeTok{albumin =} \FunctionTok{list}\NormalTok{(}\AttributeTok{fixed =}\NormalTok{ albumin }\SpecialCharTok{\textasciitilde{}}\NormalTok{ time,}
                                    \AttributeTok{random =} \SpecialCharTok{\textasciitilde{}}\NormalTok{ time),}
                     \AttributeTok{alkaline =} \FunctionTok{list}\NormalTok{(}\AttributeTok{fixed =}\NormalTok{ alkaline }\SpecialCharTok{\textasciitilde{}}\NormalTok{ time,}
                                     \AttributeTok{random =} \SpecialCharTok{\textasciitilde{}}\NormalTok{ time))}
\end{Highlighting}
\end{Shaded}

For this illustration, the outcome object contains a list with
\texttt{type} set to \texttt{surv} (for survival data) and \texttt{Y}
contain's a dataframe in wide format (one subject per row) with:
\texttt{id} the unique patient identifier; \texttt{years} the
time-to-event data; \texttt{event} the event indicator.

\begin{Shaded}
\begin{Highlighting}[]
\NormalTok{Y }\OtherTok{\textless{}{-}} \FunctionTok{list}\NormalTok{(}\AttributeTok{type =} \StringTok{"surv"}\NormalTok{,}
          \AttributeTok{Y =} \FunctionTok{unique}\NormalTok{(pbc2\_train[,}\FunctionTok{c}\NormalTok{(}\StringTok{"id"}\NormalTok{,}\StringTok{"years"}\NormalTok{,}\StringTok{"event"}\NormalTok{)]))}
\end{Highlighting}
\end{Shaded}

\hypertarget{random-forest-building}{%
\subsection{Random forest building}\label{random-forest-building}}

We build the random forest using \texttt{DynForest()} function with the
following code:

\begin{Shaded}
\begin{Highlighting}[]
\NormalTok{res\_dyn }\OtherTok{\textless{}{-}} \FunctionTok{DynForest}\NormalTok{(}\AttributeTok{timeData =}\NormalTok{ timeData\_train, }
                     \AttributeTok{fixedData =}\NormalTok{ fixedData\_train,}
                     \AttributeTok{timeVar =} \StringTok{"time"}\NormalTok{, }\AttributeTok{idVar =} \StringTok{"id"}\NormalTok{, }
                     \AttributeTok{timeVarModel =}\NormalTok{ timeVarModel, }\AttributeTok{Y =}\NormalTok{ Y,}
                     \AttributeTok{ntree =} \DecValTok{200}\NormalTok{, }\AttributeTok{mtry =} \DecValTok{3}\NormalTok{, }\AttributeTok{nodesize =} \DecValTok{2}\NormalTok{, }\AttributeTok{minsplit =} \DecValTok{3}\NormalTok{,}
                     \AttributeTok{cause =} \DecValTok{2}\NormalTok{, }\AttributeTok{ncores =} \DecValTok{7}\NormalTok{, }\AttributeTok{seed =} \DecValTok{1234}\NormalTok{)}
\end{Highlighting}
\end{Shaded}

In a survival context with multiple events, it is necessary to specify
the event of interest with the argument \texttt{cause.} We thus fixed
\texttt{cause} = 2 to specify the event of interest (i.e., the death
event). For the hyperparameters, we arbitrarily chose \texttt{mtry} = 3,
\texttt{nodesize} = 2 and \texttt{minsplit} = 3 and we will discuss this
point in section 4.8.

Overall information about the random forest can be output with the
\texttt{summary()} function as displayed below for our example:

\begin{Shaded}
\begin{Highlighting}[]
\FunctionTok{summary}\NormalTok{(res\_dyn)}
\end{Highlighting}
\end{Shaded}

\begin{verbatim}
## DynForest executed for survival (competing risk) outcome 
##  Splitting rule: Fine & Gray statistic test 
##  Out-of-bag error type: Integrated Brier Score 
##  Leaf statistic: Cumulative incidence function 
## ---------------- 
## Input 
##  Number of subjects: 208 
##  Longitudinal: 4 predictor(s) 
##  Numeric: 1 predictor(s) 
##  Factor: 2 predictor(s) 
## ---------------- 
## Tuning parameters 
##  mtry: 3 
##  nodesize: 2 
##  minsplit: 3 
##  ntree: 200 
## ---------------- 
## ---------------- 
## DynForest summary 
##  Average depth per tree: 6.62 
##  Average number of leaves per tree: 27.68 
##  Average number of subjects per leaf: 4.78 
##  Average number of events of interest per leaf: 1.95 
## ---------------- 
## Computation time 
##  Number of cores used: 7 
##  Time difference of 7.100826 mins
## ----------------
\end{verbatim}

We executed \texttt{DynForest()} function for a survival outcome with
competing events. In this mode, we use the Fine \& Gray statistic test
as the splitting rule and the cumulative incidence function (CIF) as the
leaf statistic. To build the random forest, we included 208 subjects
with 4 longitudinal (\texttt{Longitudinal}), 1 continuous
(\texttt{Numeric}) and 2 categorical (\texttt{Factor}) predictors. The
\texttt{summary()} function returns some statistics about the trees. For
instance, we have on average 4.8 subjects and 1.9 death events per leaf.
The number of subjects per leaf should always be higher than
\texttt{nodesize} hyperparameter. OOB error should be first computed
using \texttt{compute\_OOBerror()} function (see section 4.5) to be
displayed on summary output.

To further investigate the tree structure, the split details can be
output using \texttt{getTree()} function with the following code (for
tree 1):

\begin{Shaded}
\begin{Highlighting}[]
\FunctionTok{head}\NormalTok{(}\FunctionTok{getTree}\NormalTok{(}\AttributeTok{DynForest\_obj =}\NormalTok{ res\_dyn, }\AttributeTok{tree =} \DecValTok{1}\NormalTok{))}
\end{Highlighting}
\end{Shaded}

\begin{verbatim}
##           type id_node var_split feature     threshold   N Nevent depth
## 1 Longitudinal       1         3       1 -1.272629e-01 129     51     1
## 2      Numeric       2         1      NA  4.138210e+01  39     27     2
## 3 Longitudinal       3         4       1  1.459346e+02  90     24     2
## 4 Longitudinal       4         3       1  3.608271e-11   8      3     3
## 5 Longitudinal       5         2       1  5.924123e+01  31     24     3
## 6 Longitudinal       6         1       1  2.786575e-01  63     12     3
\end{verbatim}

\begin{Shaded}
\begin{Highlighting}[]
\FunctionTok{tail}\NormalTok{(}\FunctionTok{getTree}\NormalTok{(}\AttributeTok{DynForest\_obj =}\NormalTok{ res\_dyn, }\AttributeTok{tree =} \DecValTok{1}\NormalTok{))}
\end{Highlighting}
\end{Shaded}

\begin{verbatim}
##            type id_node var_split feature     threshold N Nevent depth
## 50         Leaf     174        NA      NA            NA 2      2     8
## 51 Longitudinal     175         2       1 -1.850322e-10 4      4     8
## 52         Leaf     250        NA      NA            NA 5      1     8
## 53         Leaf     251        NA      NA            NA 4      2     8
## 54         Leaf     350        NA      NA            NA 2      2     9
## 55         Leaf     351        NA      NA            NA 2      2     9
\end{verbatim}

\noindent Looking at the head of \texttt{getTree()} function output, we
see that subjects were split at node 1 (\texttt{id\_node}) using the
first random-effect (\texttt{feature} = 1) of the third
\texttt{Longitudinal} predictor (\texttt{var\_split} = 3) with
\texttt{threshold} = -0.1273. \texttt{var\_split} = 3 corresponds to
\texttt{albumin}, so subjects at node 1 with \texttt{albumin} values
below to -0.1273 are assigned in node 2, otherwise in node 3. The last
rows of random forest given by the tail of \texttt{getTree()} function
output provide the leaves descriptions. For instance, row 53, 4 subjects
are included in leaf 251, and among 2 subjects have the event of
interest.

Estimated cumulative incidence function (CIF) which in each leaf of a
tree can be displayed using \texttt{plot()} function. For instance, the
CIF of the cause of interest for leaf 251 in the tree 1 can be displayed
using the following code:

CIF of a single tree is not meant to be interpreted alone. The CIF
should be average over all trees of the random forest. For a subject,
estimated CIF over the random forest is obtained by averaging all the
tree-specific CIF of the tree-leaf where the subject belongs. This can
be done with the \texttt{plot()} function such as:

\begin{Shaded}
\begin{Highlighting}[]
\FunctionTok{plot}\NormalTok{(res\_dyn, }\AttributeTok{id =} \DecValTok{104}\NormalTok{, }\AttributeTok{max\_tree =} \DecValTok{9}\NormalTok{)}
\end{Highlighting}
\end{Shaded}

\begin{figure}

{\centering \includegraphics[width=0.8\linewidth,height=0.4\textheight]{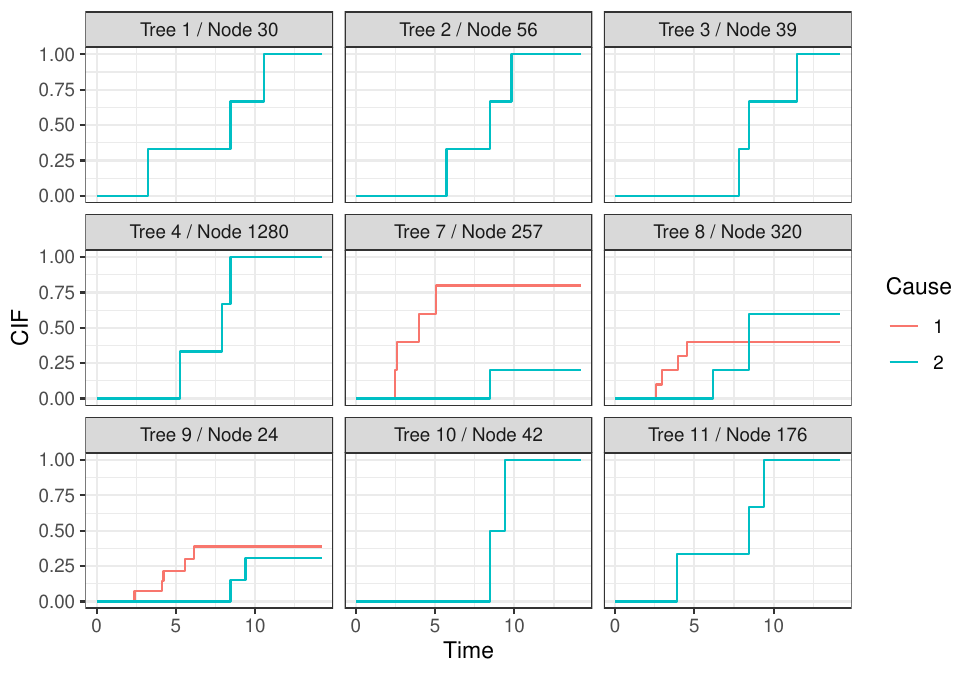} 

}

\caption{Estimated cumulative incidence functions for subject 104 over 9 trees.}\label{fig:DynForestRCIF}
\end{figure}

In this example, we display in figure \ref{fig:DynForestRCIF} for
subject 104 the tree-specific CIF for the 9 first trees where this
subject is used to grow the trees. This figure shows how the estimated
CIF can be differ across the trees and requires to be averaged as each
is calculated from information of the few subjects belonging to a leaf.

\hypertarget{out-of-bag-error}{%
\subsection{Out-Of-Bag error}\label{out-of-bag-error}}

The Out-Of-Bag error (OOB) aims at assessing the prediction abilities of
the random forest. With a survival outcome, the OOB error is evaluated
using the Integrated Brier Score (IBS) (Gerds and Schumacher 2006). It
is computed using \texttt{compute\_OOBerror()} function with an object
of class \texttt{DynForest} as main argument, such as:

\begin{Shaded}
\begin{Highlighting}[]
\NormalTok{res\_dyn\_OOB }\OtherTok{\textless{}{-}} \FunctionTok{compute\_OOBerror}\NormalTok{(}\AttributeTok{DynForest\_obj =}\NormalTok{ res\_dyn)}
\end{Highlighting}
\end{Shaded}

\texttt{compute\_OOBerror()} returns the OOB errors by individual. The
overall OOB error for the random forest is obtained by averaging the
individual specific OOB error, and can be displayed using
\texttt{print()}.

\begin{Shaded}
\begin{Highlighting}[]
\FunctionTok{print}\NormalTok{(res\_dyn\_OOB)}
\end{Highlighting}
\end{Shaded}

\begin{verbatim}
## [1] 0.1266897
\end{verbatim}

We obtain an IBS of 0.127 computed from time 0 to the maximum event
time. The time range can be modified using \texttt{IBS.min} and
\texttt{IBS.max} arguments to define the minimum and maximum,
respectively. To maximize the prediction ability of the random forest,
the hyperparameters can be tuned, that is chosen as those that minimize
the OOB error (see section 4.8).

\hypertarget{prediction-of-the-outcome}{%
\subsection{Prediction of the outcome}\label{prediction-of-the-outcome}}

The \texttt{predict()} function allows to predict the outcome for a new
subject using the trained random forest. The function requires the
individual data: time-dependent predictors in \texttt{timeData} and
time-fixed predictors in \texttt{fixedData.} For a survival outcome,
dynamic predictions can be computed by fixing a prediction time (called
landmark time, argument \texttt{t0}) from which prediction is made. In
this case, only the history of the individual up to this landmark time
(including the longitudinal and time-fixed predictors) will be used.

For the illustration, we only select the subjects still at risk at the
landmark time of 4 years. We build the dataframe for those subjects and
we predict the individual-specific CIF using \texttt{predict()} function
as follows:

\begin{Shaded}
\begin{Highlighting}[]
\NormalTok{id\_pred }\OtherTok{\textless{}{-}} \FunctionTok{unique}\NormalTok{(pbc2\_pred}\SpecialCharTok{$}\NormalTok{id[}\FunctionTok{which}\NormalTok{(pbc2\_pred}\SpecialCharTok{$}\NormalTok{years}\SpecialCharTok{\textgreater{}}\DecValTok{4}\NormalTok{)])}
\NormalTok{pbc2\_pred\_tLM }\OtherTok{\textless{}{-}}\NormalTok{ pbc2\_pred[}\FunctionTok{which}\NormalTok{(pbc2\_pred}\SpecialCharTok{$}\NormalTok{id }\SpecialCharTok{\%in\%}\NormalTok{ id\_pred),]}
\NormalTok{timeData\_pred }\OtherTok{\textless{}{-}}\NormalTok{ pbc2\_pred\_tLM[,}\FunctionTok{c}\NormalTok{(}\StringTok{"id"}\NormalTok{,}\StringTok{"time"}\NormalTok{,}
                                  \StringTok{"serBilir"}\NormalTok{,}\StringTok{"SGOT"}\NormalTok{,}
                                  \StringTok{"albumin"}\NormalTok{,}\StringTok{"alkaline"}\NormalTok{)]}
\NormalTok{fixedData\_pred }\OtherTok{\textless{}{-}} \FunctionTok{unique}\NormalTok{(pbc2\_pred\_tLM[,}\FunctionTok{c}\NormalTok{(}\StringTok{"id"}\NormalTok{,}\StringTok{"age"}\NormalTok{,}\StringTok{"drug"}\NormalTok{,}\StringTok{"sex"}\NormalTok{)])}
\NormalTok{pred\_dyn }\OtherTok{\textless{}{-}} \FunctionTok{predict}\NormalTok{(}\AttributeTok{object =}\NormalTok{ res\_dyn, }
                    \AttributeTok{timeData =}\NormalTok{ timeData\_pred, }
                    \AttributeTok{fixedData =}\NormalTok{ fixedData\_pred,}
                    \AttributeTok{idVar =} \StringTok{"id"}\NormalTok{, }\AttributeTok{timeVar =} \StringTok{"time"}\NormalTok{,}
                    \AttributeTok{t0 =} \DecValTok{4}\NormalTok{)}
\end{Highlighting}
\end{Shaded}

\begin{figure}

{\centering \includegraphics[width=0.8\linewidth,height=0.4\textheight]{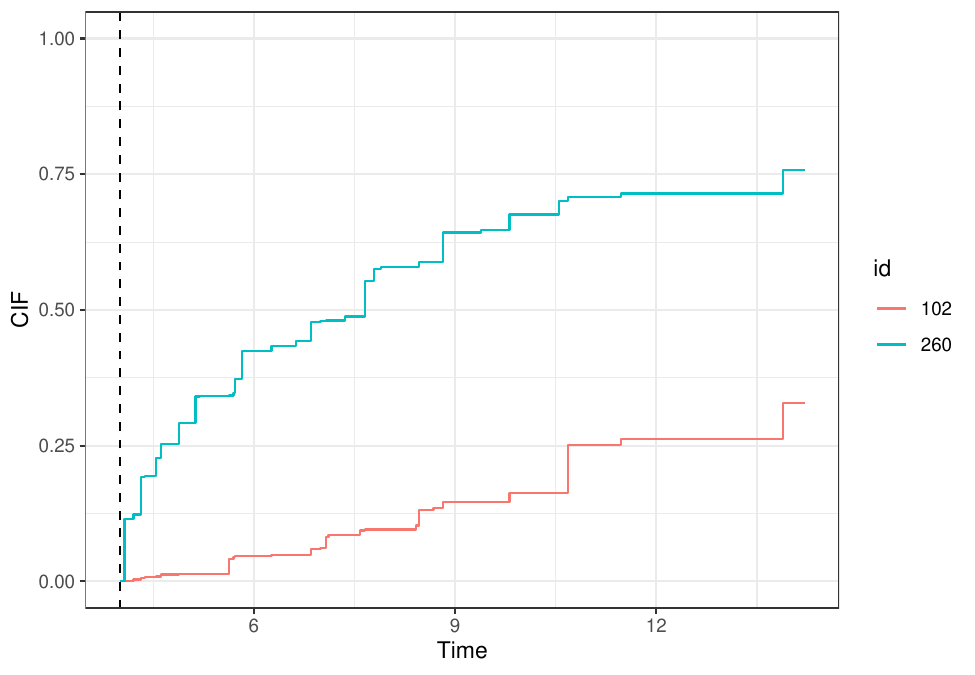} 

}

\caption{Predicted cumulative incidence function for subjects 102 and 260 from landmark time of 4 years (represented by the dashed vertical line)}\label{fig:DynForestRpredCIF}
\end{figure}

The \texttt{predict()} function provides several elements as described
in section 3.2. In addition, the \texttt{plot()} function can be used to
display the CIF of the outcome (here death before transplantation) for
subjects indicated with argument \texttt{id}. For instance, we compute
the CIF for subjects 102 and 260 with the following code and display
them in figure \ref{fig:DynForestRpredCIF}.

The first year after the landmark time (at 4 years), we observe a rapid
increase of the risk of death for subject 260 compared to subject 102.
We also notice that after 10 years from landmark time, subject 260 has a
probability of death almost three times higher that the one of subject
102.

\hypertarget{predictiveness-of-the-variables}{%
\subsection{Predictiveness of the
variables}\label{predictiveness-of-the-variables}}

\hypertarget{variable-importance-1}{%
\subsubsection{Variable importance}\label{variable-importance-1}}

The main objective of the random forest is to predict an outcome. But
usually, we are interested in identifying which predictors are the most
predictive. The VIMP statistic (Hemant Ishwaran et al. 2008) can be
computed using \texttt{compute\_VIMP()} function. This function returns
the VIMP statistic for each predictor with \texttt{\$Importance}
element. These results can also be displayed using \texttt{plot()}
function, either in absolute value by default or in percentage with
\texttt{PCT} argument set to \texttt{TRUE}.

\begin{Shaded}
\begin{Highlighting}[]
\NormalTok{res\_dyn\_VIMP }\OtherTok{\textless{}{-}} \FunctionTok{compute\_VIMP}\NormalTok{(}\AttributeTok{DynForest\_obj =}\NormalTok{ res\_dyn, }\AttributeTok{seed =} \DecValTok{123}\NormalTok{)}
\end{Highlighting}
\end{Shaded}

\begin{Shaded}
\begin{Highlighting}[]
\NormalTok{p1 }\OtherTok{\textless{}{-}} \FunctionTok{plot}\NormalTok{(res\_dyn\_VIMP, }\AttributeTok{PCT =} \ConstantTok{TRUE}\NormalTok{)}
\end{Highlighting}
\end{Shaded}

\begin{figure}

{\centering \includegraphics[width=0.8\linewidth,height=0.4\textheight]{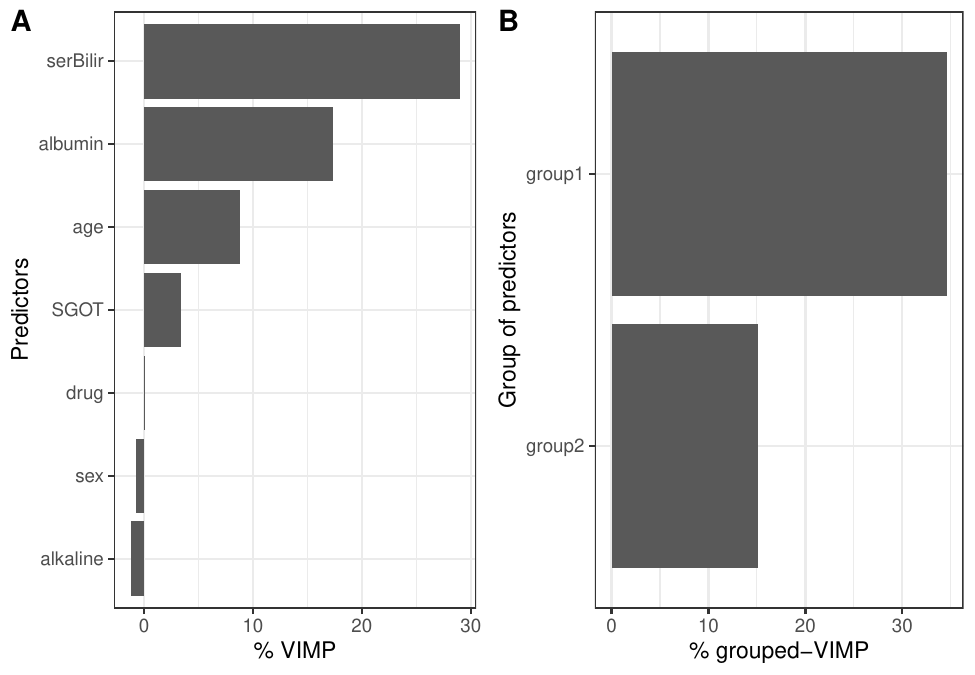} 

}

\caption{(A) VIMP statistic and (B) grouped-VIMP statistic displayed as a percentage of loss in OOB error of prediction. group1 includes serBilir and SGOT; group2 includes albumin and alkaline.}\label{fig:DynForestRVIMPgVIMP}
\end{figure}

The VIMP results are displayed in figure \ref{fig:DynForestRVIMPgVIMP}A.
The most predictive variables are \texttt{serBilir}, \texttt{albumin}
and \texttt{age} with the largest VIMP percentage. By removing the
association between \texttt{serBilir} and the event, the OOB error was
increased by 30\%.

In the case of correlated predictors, the predictors can be regrouped
into dimensions and the VIMP can be computed at the dimension group
level with the gVIMP statistic. Permutation is done for each variable of
the group simultaneously. The gVIMP is computed with the
\texttt{compute\_gVIMP()} function in which the \texttt{group} argument
defines the group of predictors as a list. For instance, with two groups
of predictors (named group1 and group2), the gVIMP statistic is computed
using the following code:

\begin{Shaded}
\begin{Highlighting}[]
\NormalTok{group }\OtherTok{\textless{}{-}} \FunctionTok{list}\NormalTok{(}\AttributeTok{group1 =} \FunctionTok{c}\NormalTok{(}\StringTok{"serBilir"}\NormalTok{,}\StringTok{"SGOT"}\NormalTok{),}
              \AttributeTok{group2 =} \FunctionTok{c}\NormalTok{(}\StringTok{"albumin"}\NormalTok{,}\StringTok{"alkaline"}\NormalTok{))}
\NormalTok{res\_dyn\_gVIMP }\OtherTok{\textless{}{-}} \FunctionTok{compute\_gVIMP}\NormalTok{(}\AttributeTok{DynForest\_obj =}\NormalTok{ res\_dyn,}
                               \AttributeTok{group =}\NormalTok{ group, }\AttributeTok{seed =} \DecValTok{123}\NormalTok{)}
\end{Highlighting}
\end{Shaded}

\begin{Shaded}
\begin{Highlighting}[]
\NormalTok{p2 }\OtherTok{\textless{}{-}} \FunctionTok{plot}\NormalTok{(res\_dyn\_gVIMP, }\AttributeTok{PCT =} \ConstantTok{TRUE}\NormalTok{)}
\end{Highlighting}
\end{Shaded}

Similar to VIMP statistic, the gVIMP results can be displayed using
\texttt{plot()} function. The figure \ref{fig:DynForestRVIMPgVIMP}B
shows that group1 has the highest gVIMP percentage with 34\%.

\begin{Shaded}
\begin{Highlighting}[]
\FunctionTok{plot\_grid}\NormalTok{(p1, p2, }\AttributeTok{labels =} \FunctionTok{c}\NormalTok{(}\StringTok{"A"}\NormalTok{, }\StringTok{"B"}\NormalTok{))}
\end{Highlighting}
\end{Shaded}

To compute the gVIMP statistic, the groups can be defined regardless of
the number of predictors. However, the comparison between the groups may
be harder when group sizes are very different.

\hypertarget{minimal-depth-1}{%
\subsubsection{Minimal depth}\label{minimal-depth-1}}

To go further into the understanding of the tree building process, the
\texttt{var\_depth()} function extracts information about the average
minimal depth by feature (\texttt{\$min\_depth}), the minimal depth for
each feature and each tree (\texttt{\$var\_node\_depth}), the number of
times that the feature is used for splitting for each feature and each
tree (\texttt{\$var\_count}).

Using an object from \texttt{var\_depth()} function, \texttt{plot()}
function allows to plot the distribution of the average minimal depth
across the trees. \texttt{plot\_level} argument defines how the average
minimal depth is plotted, by predictor or feature.

\begin{figure}

{\centering \includegraphics[width=0.8\linewidth,height=0.4\textheight]{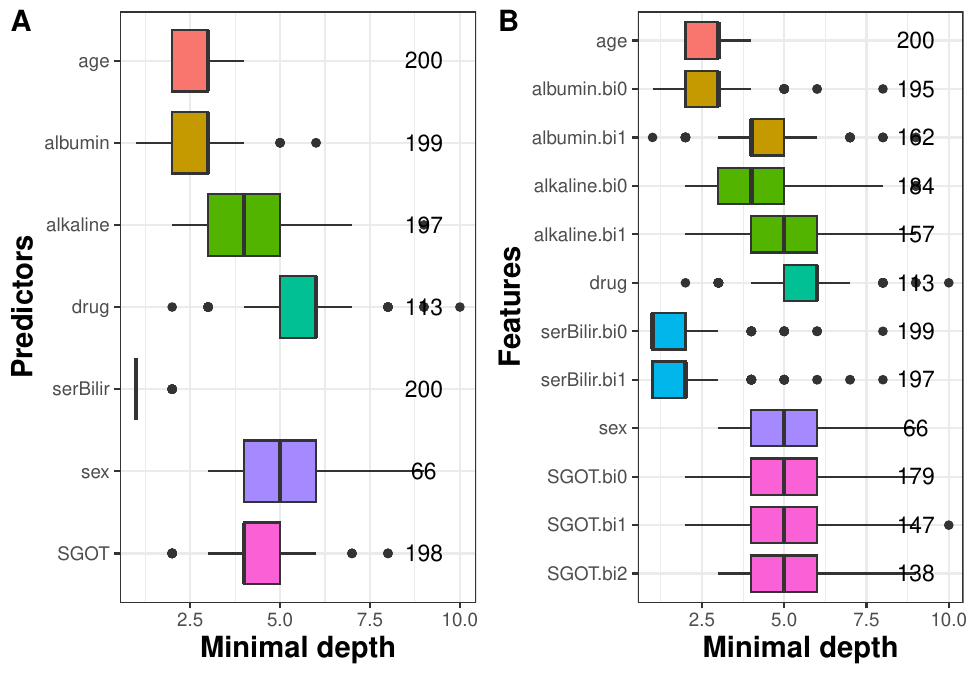} 

}

\caption{Average minimal depth level by predictor (A) and feature (B).}\label{fig:DynForestRmindepth}
\end{figure}

The distribution of the minimal depth level is displayed in figure
\ref{fig:DynForestRmindepth} by predictor and feature. Note that the
minimal depth level should always be interpreted with the number of
trees where the predictor/feature is found. Indeed, to accurately
appreciate the importance of a variable minimal depth, the variable has
to be systematically part of the candidates at each node. This is why we
strongly advice to compute the minimal depth on random forest with
\texttt{mtry} hyperparameter chosen at its maximum (as done below).

\begin{Shaded}
\begin{Highlighting}[]
\NormalTok{res\_dyn\_max }\OtherTok{\textless{}{-}} \FunctionTok{DynForest}\NormalTok{(}\AttributeTok{timeData =}\NormalTok{ timeData\_train, }
                         \AttributeTok{fixedData =}\NormalTok{ fixedData\_train,}
                         \AttributeTok{timeVar =} \StringTok{"time"}\NormalTok{, }\AttributeTok{idVar =} \StringTok{"id"}\NormalTok{, }
                         \AttributeTok{timeVarModel =}\NormalTok{ timeVarModel, }\AttributeTok{Y =}\NormalTok{ Y,}
                         \AttributeTok{ntree =} \DecValTok{200}\NormalTok{, }\AttributeTok{mtry =} \DecValTok{7}\NormalTok{, }\AttributeTok{nodesize =} \DecValTok{2}\NormalTok{, }\AttributeTok{minsplit =} \DecValTok{3}\NormalTok{, }
                         \AttributeTok{cause =} \DecValTok{2}\NormalTok{, }\AttributeTok{ncores =} \DecValTok{7}\NormalTok{, }\AttributeTok{seed =} \DecValTok{1234}\NormalTok{)}
\end{Highlighting}
\end{Shaded}

\begin{Shaded}
\begin{Highlighting}[]
\NormalTok{depth\_dyn }\OtherTok{\textless{}{-}} \FunctionTok{var\_depth}\NormalTok{(}\AttributeTok{DynForest\_obj =}\NormalTok{ res\_dyn\_max)}
\NormalTok{p1 }\OtherTok{\textless{}{-}} \FunctionTok{plot}\NormalTok{(depth\_dyn, }\AttributeTok{plot\_level =} \StringTok{"predictor"}\NormalTok{)}
\end{Highlighting}
\end{Shaded}

\begin{Shaded}
\begin{Highlighting}[]
\NormalTok{p2 }\OtherTok{\textless{}{-}} \FunctionTok{plot}\NormalTok{(depth\_dyn, }\AttributeTok{plot\_level =} \StringTok{"feature"}\NormalTok{)}
\end{Highlighting}
\end{Shaded}

\begin{Shaded}
\begin{Highlighting}[]
\FunctionTok{plot\_grid}\NormalTok{(p1, p2, }\AttributeTok{labels =} \FunctionTok{c}\NormalTok{(}\StringTok{"A"}\NormalTok{, }\StringTok{"B"}\NormalTok{))}
\end{Highlighting}
\end{Shaded}

In our example, we ran a random forest with mtry hyperparameter set to
its maximum (i.e., \texttt{mtry} = 7) and we computed the minimal depth
on this random forest. We observe that \texttt{serBilir},
\texttt{albumin} and \texttt{age} have the lowest minimal depth,
indicating these predictors are used to split the subjects at early
stages in 200 out of 200 trees, i.e., 100\% for \texttt{serBilir},
\texttt{age} and in 199 out of 200 for \texttt{albumin} (figure
\ref{fig:DynForestRmindepth}A). The minimal depth level by feature
(figure \ref{fig:DynForestRmindepth}B) provides more advanced details
about the tree building process. For instance, we can see that the
random-effects of \texttt{serBilir} (indicating by bi0 and bi1 in the
graph) are the earliest features used on 199 and 197 out of 200 trees,
respectively.

\hypertarget{guidelines-to-tune-the-hyperparameters}{%
\subsection{Guidelines to tune the
hyperparameters}\label{guidelines-to-tune-the-hyperparameters}}

The predictive performance of the random forest strongly depends on the
hyperparameters \texttt{mtry}, \texttt{nodesize} and \texttt{minsplit.}
They should therefore be chosen thoroughly. \texttt{nodesize} and
\texttt{minsplit} hyperparameters control the tree depth. The trees need
to be deep enough to ensure that the predictions are accurate. By
default, we fixed \texttt{nodesize} and \texttt{minsplit} at the
minimum, that is \texttt{nodesize} = 1 and \texttt{minsplit} = 2.
However, with a large number of individuals, the tree depth could be
slightly decreased by increasing these hyperparameters in order to
reduce the computation time.

\texttt{mtry} hyperparameter defines the number of predictors randomly
drawn at each node. By default, we chose \texttt{mtry} equal to the
square root of the number of predictors as usually recommended (Bernard,
Heutte, and Adam 2009). However, this hyperparameter should be carefully
tuned with possible values between 1 and the number of predictors.
Indeed, the predictive performance of the random forest is highly
related to this hyperparameter.

\begin{figure}

{\centering \includegraphics[width=0.8\linewidth,height=0.4\textheight]{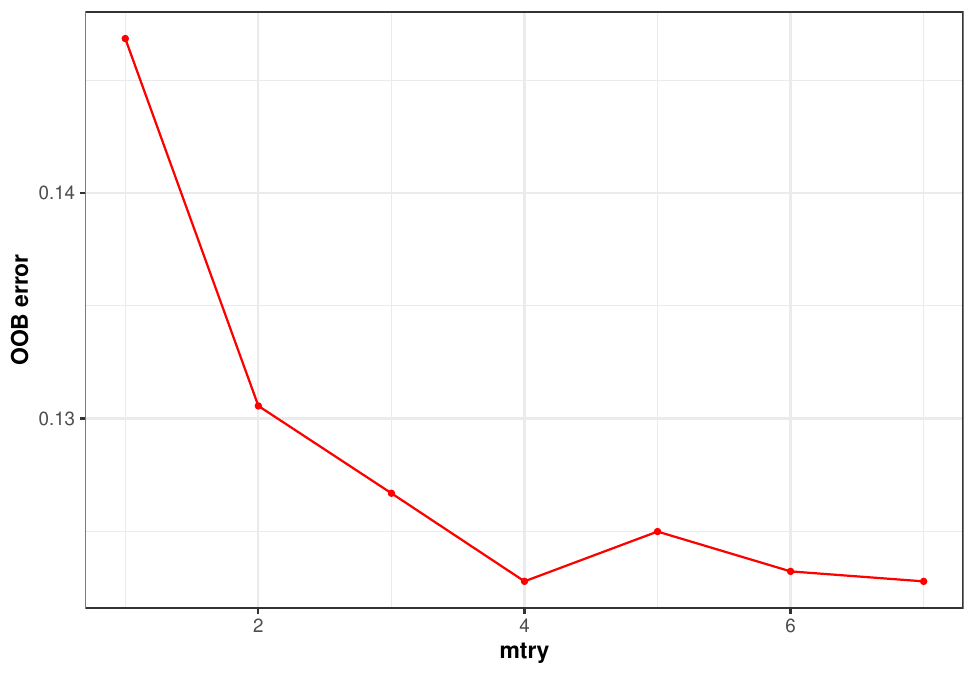} 

}

\caption{OOB error according to mtry hyperparameter. The optimal value was found for the maximum value mtry = 7.}\label{fig:DynForestRmtrytuned}
\end{figure}

In the illustration, we tuned \texttt{mtry} for every possible values (1
to 7). Figure \ref{fig:DynForestRmtrytuned} displays the OOB error
according to \texttt{mtry} hyperparameter.

We can see on this figure large OOB error difference according to
\texttt{mtry} hyperparameter. In particular, we observe the worst
predictive performance for lower values, then similar results with
values from 4 to 7. The optimal value (i.e., with the lowest OOB error)
was found with the maximum value \texttt{mtry} = 7. This graph reflects
how crucial it is to carefully tune this hyperparameter.

\hypertarget{how-to-use-dynforest-r-package-with-a-categorical-outcome}{%
\section{How to use DynForest R package with a categorical
outcome?}\label{how-to-use-dynforest-r-package-with-a-categorical-outcome}}

In this section, we use \textbf{DynForest} in a classification
perspective using \texttt{pbc2} data. For the illustration purpose, we
want to predict the death between 4 and 10 years on subjects still at
risk at 4 years from the repeated data up to 4 years. Note that this is
only for illustrative purpose as this technique does not handle
censoring correctly.

\hypertarget{data-management-1}{%
\subsection{Data management}\label{data-management-1}}

For the illustration, we select patients still at risk at 4 years and we
recode the \texttt{event} variable with \texttt{event} = 1 for subjects
who died during between 4 years and 10 years, whereas subjects with
transplantation were recoded \texttt{event} = 0, as the subjects still
alive. We split the subjects into two datasets: (i) one dataset to train
the random forest using \(2/3\) of patients; (ii) one dataset to predict
on the other \(1/3\) of patients.

We use the same strategy as in the survival context (section 4) to build
the random forest, with the same predictors and the same association for
time-dependent predictors.

\begin{Shaded}
\begin{Highlighting}[]
\NormalTok{timeData\_train }\OtherTok{\textless{}{-}}\NormalTok{ pbc2\_train[,}\FunctionTok{c}\NormalTok{(}\StringTok{"id"}\NormalTok{,}\StringTok{"time"}\NormalTok{,}
                                \StringTok{"serBilir"}\NormalTok{,}\StringTok{"SGOT"}\NormalTok{,}
                                \StringTok{"albumin"}\NormalTok{,}\StringTok{"alkaline"}\NormalTok{)]}
\NormalTok{timeVarModel }\OtherTok{\textless{}{-}} \FunctionTok{list}\NormalTok{(}\AttributeTok{serBilir =} \FunctionTok{list}\NormalTok{(}\AttributeTok{fixed =}\NormalTok{ serBilir }\SpecialCharTok{\textasciitilde{}}\NormalTok{ time,}
                                     \AttributeTok{random =} \SpecialCharTok{\textasciitilde{}}\NormalTok{ time),}
                     \AttributeTok{SGOT =} \FunctionTok{list}\NormalTok{(}\AttributeTok{fixed =}\NormalTok{ SGOT }\SpecialCharTok{\textasciitilde{}}\NormalTok{ time }\SpecialCharTok{+} \FunctionTok{I}\NormalTok{(time}\SpecialCharTok{\^{}}\DecValTok{2}\NormalTok{),}
                                 \AttributeTok{random =} \SpecialCharTok{\textasciitilde{}}\NormalTok{ time }\SpecialCharTok{+} \FunctionTok{I}\NormalTok{(time}\SpecialCharTok{\^{}}\DecValTok{2}\NormalTok{)),}
                     \AttributeTok{albumin =} \FunctionTok{list}\NormalTok{(}\AttributeTok{fixed =}\NormalTok{ albumin }\SpecialCharTok{\textasciitilde{}}\NormalTok{ time,}
                                    \AttributeTok{random =} \SpecialCharTok{\textasciitilde{}}\NormalTok{ time),}
                     \AttributeTok{alkaline =} \FunctionTok{list}\NormalTok{(}\AttributeTok{fixed =}\NormalTok{ alkaline }\SpecialCharTok{\textasciitilde{}}\NormalTok{ time,}
                                     \AttributeTok{random =} \SpecialCharTok{\textasciitilde{}}\NormalTok{ time))}
\NormalTok{fixedData\_train }\OtherTok{\textless{}{-}} \FunctionTok{unique}\NormalTok{(pbc2\_train[,}\FunctionTok{c}\NormalTok{(}\StringTok{"id"}\NormalTok{,}\StringTok{"age"}\NormalTok{,}\StringTok{"drug"}\NormalTok{,}\StringTok{"sex"}\NormalTok{)])}
\end{Highlighting}
\end{Shaded}

With a categorical outcome, the definition of the output object is
slightly different. We should specify \texttt{type}=``factor'' to define
the outcome as categorical, and the dataframe in \texttt{Y} should
contain only 2 columns, the variable identifier \texttt{id} and the
outcome \texttt{event}.

\begin{Shaded}
\begin{Highlighting}[]
\NormalTok{Y }\OtherTok{\textless{}{-}} \FunctionTok{list}\NormalTok{(}\AttributeTok{type =} \StringTok{"factor"}\NormalTok{,}
          \AttributeTok{Y =} \FunctionTok{unique}\NormalTok{(pbc2\_train[,}\FunctionTok{c}\NormalTok{(}\StringTok{"id"}\NormalTok{,}\StringTok{"event"}\NormalTok{)]))}
\end{Highlighting}
\end{Shaded}

\hypertarget{the-random-forest-building}{%
\subsection{The random forest
building}\label{the-random-forest-building}}

We executed \texttt{DynForest()} function to build the random forest
with hyperparameters \texttt{mtry} = 7 and \texttt{nodesize} = 2 as
follows:

\begin{Shaded}
\begin{Highlighting}[]
\NormalTok{res\_dyn }\OtherTok{\textless{}{-}} \FunctionTok{DynForest}\NormalTok{(}\AttributeTok{timeData =}\NormalTok{ timeData\_train, }
                     \AttributeTok{fixedData =}\NormalTok{ fixedData\_train,}
                     \AttributeTok{timeVar =} \StringTok{"time"}\NormalTok{, }\AttributeTok{idVar =} \StringTok{"id"}\NormalTok{, }
                     \AttributeTok{timeVarModel =}\NormalTok{ timeVarModel,}
                     \AttributeTok{mtry =} \DecValTok{7}\NormalTok{, }\AttributeTok{nodesize =} \DecValTok{2}\NormalTok{, }
                     \AttributeTok{Y =}\NormalTok{ Y, }\AttributeTok{ncores =} \DecValTok{7}\NormalTok{, }\AttributeTok{seed =} \DecValTok{1234}\NormalTok{)}
\end{Highlighting}
\end{Shaded}

\hypertarget{out-of-bag-error-1}{%
\subsection{Out-Of-Bag error}\label{out-of-bag-error-1}}

With a categorical outcome, the OOB prediction error is evaluated using
a missclassification criterion. This criterion can be computed with
\texttt{compute\_OOBerror()} function and the results of the random
forest can be displayed using \texttt{summary()}:

\begin{Shaded}
\begin{Highlighting}[]
\NormalTok{res\_dyn\_OOB }\OtherTok{\textless{}{-}} \FunctionTok{compute\_OOBerror}\NormalTok{(}\AttributeTok{DynForest\_obj =}\NormalTok{ res\_dyn)}
\end{Highlighting}
\end{Shaded}

\begin{Shaded}
\begin{Highlighting}[]
\FunctionTok{summary}\NormalTok{(res\_dyn\_OOB)}
\end{Highlighting}
\end{Shaded}

\begin{verbatim}
## DynForest executed for categorical outcome 
##  Splitting rule: Minimize weighted within-group Shannon entropy 
##  Out-of-bag error type: Missclassification 
##  Leaf statistic: Majority vote 
## ---------------- 
## Input 
##  Number of subjects: 150 
##  Longitudinal: 4 predictor(s) 
##  Numeric: 1 predictor(s) 
##  Factor: 2 predictor(s) 
## ---------------- 
## Tuning parameters 
##  mtry: 7 
##  nodesize: 2 
##  ntree: 200 
## ---------------- 
## ---------------- 
## DynForest summary 
##  Average depth per tree: 5.89 
##  Average number of leaves per tree: 16.8 
##  Average number of subjects per leaf: 5.73 
## ---------------- 
## Out-of-bag error based on Missclassification 
##  Out-of-bag error: 0.24 
## ---------------- 
## Computation time 
##  Number of cores used: 7 
##  Time difference of 6.452599 mins
## ----------------
\end{verbatim}

In this illustration, we built the random forest using 150 subjects
because we only kept the subjects still at risk at landmark time at 4
years and split the dataset in \(2/3\) for training and \(1/3\) for
testing. We have on average 5.7 subjects per leaf, and the average depth
level per tree is 5.9. This random forest predicted the wrong outcome
for 24\% of the subjects. The random forest performances can be
optimized by choosing the \texttt{mtry} and \texttt{nodesize}
hyperparameters that minimized the OOB missclassification.

\hypertarget{prediction-of-the-outcome-1}{%
\subsection{Prediction of the
outcome}\label{prediction-of-the-outcome-1}}

We can predict the probability of death between 4 and 10 years on
subjects still at risk at landmark time at 4 years. In classification
mode, the predictions are performed using the majority vote. The
prediction over the trees is thus a category of the outcome along with
the proportion of the trees that lead to this category. Predictions are
computed using \texttt{predict()} function, then a dataframe can be
easily built from the returning object to get the prediction and
probability of the outcome for each subject:

\begin{Shaded}
\begin{Highlighting}[]
\NormalTok{timeData\_pred }\OtherTok{\textless{}{-}}\NormalTok{ pbc2\_pred[,}\FunctionTok{c}\NormalTok{(}\StringTok{"id"}\NormalTok{,}\StringTok{"time"}\NormalTok{,}
                              \StringTok{"serBilir"}\NormalTok{,}\StringTok{"SGOT"}\NormalTok{,}
                              \StringTok{"albumin"}\NormalTok{,}\StringTok{"alkaline"}\NormalTok{)]}
\NormalTok{fixedData\_pred }\OtherTok{\textless{}{-}} \FunctionTok{unique}\NormalTok{(pbc2\_pred[,}\FunctionTok{c}\NormalTok{(}\StringTok{"id"}\NormalTok{,}\StringTok{"age"}\NormalTok{,}\StringTok{"drug"}\NormalTok{,}\StringTok{"sex"}\NormalTok{)])}
\NormalTok{pred\_dyn }\OtherTok{\textless{}{-}} \FunctionTok{predict}\NormalTok{(}\AttributeTok{object =}\NormalTok{ res\_dyn,}
                    \AttributeTok{timeData =}\NormalTok{ timeData\_pred, }
                    \AttributeTok{fixedData =}\NormalTok{ fixedData\_pred,}
                    \AttributeTok{idVar =} \StringTok{"id"}\NormalTok{, }\AttributeTok{timeVar =} \StringTok{"time"}\NormalTok{,}
                    \AttributeTok{t0 =} \DecValTok{4}\NormalTok{)}
\end{Highlighting}
\end{Shaded}

\begin{Shaded}
\begin{Highlighting}[]
\FunctionTok{head}\NormalTok{(}\FunctionTok{data.frame}\NormalTok{(}\AttributeTok{pred =}\NormalTok{ pred\_dyn}\SpecialCharTok{$}\NormalTok{pred\_indiv, }
                \AttributeTok{proba =}\NormalTok{ pred\_dyn}\SpecialCharTok{$}\NormalTok{pred\_indiv\_proba))}
\end{Highlighting}
\end{Shaded}

\begin{verbatim}
##     pred proba
## 101    0 0.945
## 104    0 0.790
## 106    1 0.605
## 108    0 0.945
## 112    1 0.575
## 114    0 0.645
\end{verbatim}

As shown in this example, some predictions are made with varying
confidence from 57.5\% for subject 112 to 94.5\% for subject 101. We
predict for instance no event for subject 101 with a probability of
94.5\% and an event for subject 106 with a probability of 60.5\%.

\hypertarget{predictiveness-variables}{%
\subsection{Predictiveness variables}\label{predictiveness-variables}}

\hypertarget{variable-importance-2}{%
\subsubsection{Variable importance}\label{variable-importance-2}}

The most predictive variables can be identified using
\texttt{compute\_VIMP()} and displayed using \texttt{plot()} function as
follows:

\begin{Shaded}
\begin{Highlighting}[]
\NormalTok{res\_dyn\_VIMP }\OtherTok{\textless{}{-}} \FunctionTok{compute\_VIMP}\NormalTok{(}\AttributeTok{DynForest\_obj =}\NormalTok{ res\_dyn\_OOB, }\AttributeTok{seed =} \DecValTok{123}\NormalTok{)}
\FunctionTok{plot}\NormalTok{(res\_dyn\_VIMP, }\AttributeTok{PCT =} \ConstantTok{TRUE}\NormalTok{)}
\end{Highlighting}
\end{Shaded}

Again, we found that the most predictive variable is \texttt{serBilir}.
When perturbating \texttt{serBilir}, the OOB prediction error was
increased by 15\%.

\hypertarget{minimal-depth-2}{%
\subsubsection{Minimal depth}\label{minimal-depth-2}}

\begin{figure}

{\centering \includegraphics[width=0.8\linewidth,height=0.4\textheight]{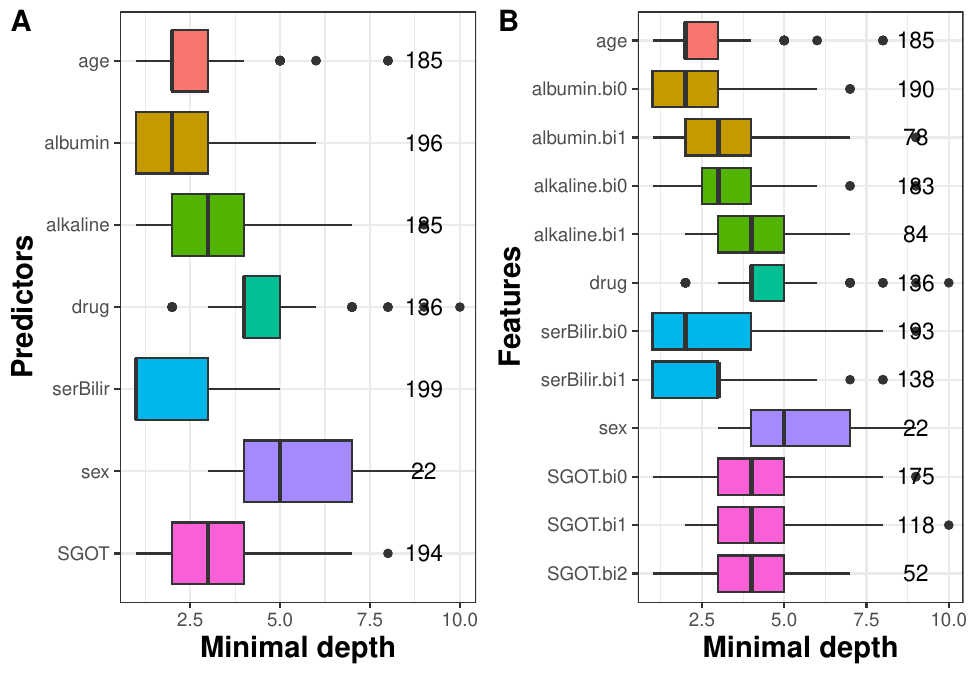} 

}

\caption{Average minimal depth by predictor (A) and feature (B).}\label{fig:DynForestRfactormindepth}
\end{figure}

The minimal depth is computed using \texttt{var\_depth()} function and
is displayed at predictor and feature levels using \texttt{plot()}
function. The results are displayed in figure
\ref{fig:DynForestRfactormindepth} using the random forest with maximal
\texttt{mtry} hyperparameter value (i.e., \texttt{mtry} = 7) for a
better understanding.

\begin{Shaded}
\begin{Highlighting}[]
\NormalTok{depth\_dyn }\OtherTok{\textless{}{-}} \FunctionTok{var\_depth}\NormalTok{(}\AttributeTok{DynForest\_obj =}\NormalTok{ res\_dyn\_OOB)}
\NormalTok{p1 }\OtherTok{\textless{}{-}} \FunctionTok{plot}\NormalTok{(depth\_dyn, }\AttributeTok{plot\_level =} \StringTok{"predictor"}\NormalTok{)}
\end{Highlighting}
\end{Shaded}

\begin{Shaded}
\begin{Highlighting}[]
\NormalTok{p2 }\OtherTok{\textless{}{-}} \FunctionTok{plot}\NormalTok{(depth\_dyn, }\AttributeTok{plot\_level =} \StringTok{"feature"}\NormalTok{)}
\end{Highlighting}
\end{Shaded}

\begin{Shaded}
\begin{Highlighting}[]
\FunctionTok{plot\_grid}\NormalTok{(p1, p2, }\AttributeTok{labels =} \FunctionTok{c}\NormalTok{(}\StringTok{"A"}\NormalTok{, }\StringTok{"B"}\NormalTok{))}
\end{Highlighting}
\end{Shaded}

We observe that \texttt{serBilir} and \texttt{albumin} have the lowest
minimal depth and are used to split the subjects in almost all the trees
(199 and 196 out of 200 trees, respectively) (figure
\ref{fig:DynForestRfactormindepth}A). Figure
\ref{fig:DynForestRfactormindepth}B provides further results. In
particular, this graph shows that the random intercept (indicated by
bi0) of \texttt{serBilir} and \texttt{albumin} are the earliest
predictors used to split the subjects and are present in 193 and 190 out
of 200 trees, respectively.

\hypertarget{how-to-use-dynforest-r-package-with-a-continuous-outcome}{%
\section{How to use DynForest R package with a continuous
outcome?}\label{how-to-use-dynforest-r-package-with-a-continuous-outcome}}

In this section, we present an illustration of \textbf{DynForest} with a
continuous outcome. \textbf{DynForest} was used on a simulated dataset
with 200 subjects and 10 predictors (6 time-dependent and 4 time-fixed
predictors). The 6 longitudinal predictors were generated using a linear
mixed model with linear trajectory according to time. We considered 6
measurements by subject (at baseline and then randomly drawn around
theoretical annual visits up to 5 years). Then, we generated the
continuous outcome using a linear regression with the random intercept
of marker 1 and random slope of marker 2 as linear predictors. We
generated two datasets (\texttt{data\_simu1} and \texttt{data\_simu2}),
one for each step (training and prediction). These datasets are
available in the \textbf{DynForest} package.

The aim of this illustration is to predict the continuous outcome using
time-dependent and time-fixed predictors.

\hypertarget{data-management-2}{%
\subsection{Data management}\label{data-management-2}}

First of all, we load the data and we build the mandatory objects needed
to execute \texttt{DynForest()} function that are
\texttt{timeData\_train} for time-dependent predictors and
\texttt{fixedData\_train} for time-fixed predictors. We specify the
model for the longitudinal predictors in \texttt{timeVarModel} object.
We considered linear trajectories over time for the 6 longitudinal
predictors.

To define the \texttt{Y} object for a continuous outcome, the
\texttt{type} argument should be set to \texttt{numeric} to run the
random forest in regression mode. The dataframe \texttt{Y} should
include two columns with the unique identifier \texttt{id} and the
continuous outcome, here \texttt{Y\_res}.

\hypertarget{the-random-forest-building-1}{%
\subsection{The random forest
building}\label{the-random-forest-building-1}}

To build the random forest, we chose default hyperparameters (i.e.,
\texttt{ntree} = 200 and \texttt{nodesize} = 1), except for
\texttt{mtry} which was fixed at its maximum (i.e., \texttt{mtry} = 10).
We ran \texttt{DynForest()} function with the following code:

\hypertarget{out-of-bag-error-2}{%
\subsection{Out-Of-Bag error}\label{out-of-bag-error-2}}

For continuous outcome, the OOB prediction error is evaluated using the
mean square error (MSE). We used \texttt{compute\_OOBerror()} function
to compute the OOB prediction error and we provided overall results with
\texttt{summary()} function as shown below:

\begin{Shaded}
\begin{Highlighting}[]
\FunctionTok{summary}\NormalTok{(res\_dyn\_OOB)}
\end{Highlighting}
\end{Shaded}

\begin{verbatim}
## DynForest executed for continuous outcome 
##  Splitting rule: Minimize weighted within-group variance 
##  Out-of-bag error type: Mean square error 
##  Leaf statistic: Mean 
## ---------------- 
## Input 
##  Number of subjects: 200 
##  Longitudinal: 6 predictor(s) 
##  Numeric: 2 predictor(s) 
##  Factor: 2 predictor(s) 
## ---------------- 
## Tuning parameters 
##  mtry: 10 
##  nodesize: 1 
##  ntree: 200 
## ---------------- 
## ---------------- 
## DynForest summary 
##  Average depth per tree: 9.06 
##  Average number of leaves per tree: 126.47 
##  Average number of subjects per leaf: 1 
## ---------------- 
## Out-of-bag error based on Mean square error 
##  Out-of-bag error: 4.3713 
## ---------------- 
## Computation time 
##  Number of cores used: 7 
##  Time difference of 16.60685 mins
## ----------------
\end{verbatim}

The random forest was executed in regression mode (for a continuous
outcome). The splitting rule aimed to minimize the weighted within-group
variance. We built the random forest using 200 subjects and 10
predictors (6 time-dependent and 4 time-fixed predictors) with
hyperparameters \texttt{ntree} = 200, \texttt{mtry} = 10 and
\texttt{nodesize} = 1. As we can see, \texttt{nodesize} = 1 leads to
deeper trees (the average depth by tree is 9.1) and a single subject by
leaf. We obtained 4.4 for the MSE. This quantity can be minimized by
tuning hyperparameters \texttt{mtry} and \texttt{nodesize}.

\hypertarget{prediction-of-the-outcome-2}{%
\subsection{Prediction of the
outcome}\label{prediction-of-the-outcome-2}}

In regression mode, the tree and leaf-specific means are averaged across
the trees to get a unique prediction over the random forest.
\texttt{predict()} function provides the individual predictions. We
first created the \texttt{timeData} and \texttt{fixedData} from the
testing sample \texttt{data\_simu2}. We then predicted the continuous
outcome by running \texttt{predict()} function:

Individual predictions can be extracted using \texttt{print()} function:

\begin{Shaded}
\begin{Highlighting}[]
\FunctionTok{head}\NormalTok{(}\FunctionTok{print}\NormalTok{(pred\_dyn))}
\end{Highlighting}
\end{Shaded}

\begin{verbatim}
##          1          2          3          4          5          6 
##  5.2184031 -1.2786887  0.8591368  1.5115312  5.2984117  7.9073981
\end{verbatim}

For instance, we predicted 5.22 for subject 1, -1.28 for subject 2 and
0.86 for subject 3.

\hypertarget{predictiveness-of-the-variables-1}{%
\subsection{Predictiveness of the
variables}\label{predictiveness-of-the-variables-1}}

In this illustration, we want to evaluate if \textbf{DynForest} can
identify the true predictors (i.e., random intercept of marker1 and
random slope of marker2). To do this, we used the minimal depth which
allows to understand the random forest at the feature level.

\begin{figure}

{\centering \includegraphics[width=0.8\linewidth,height=0.4\textheight]{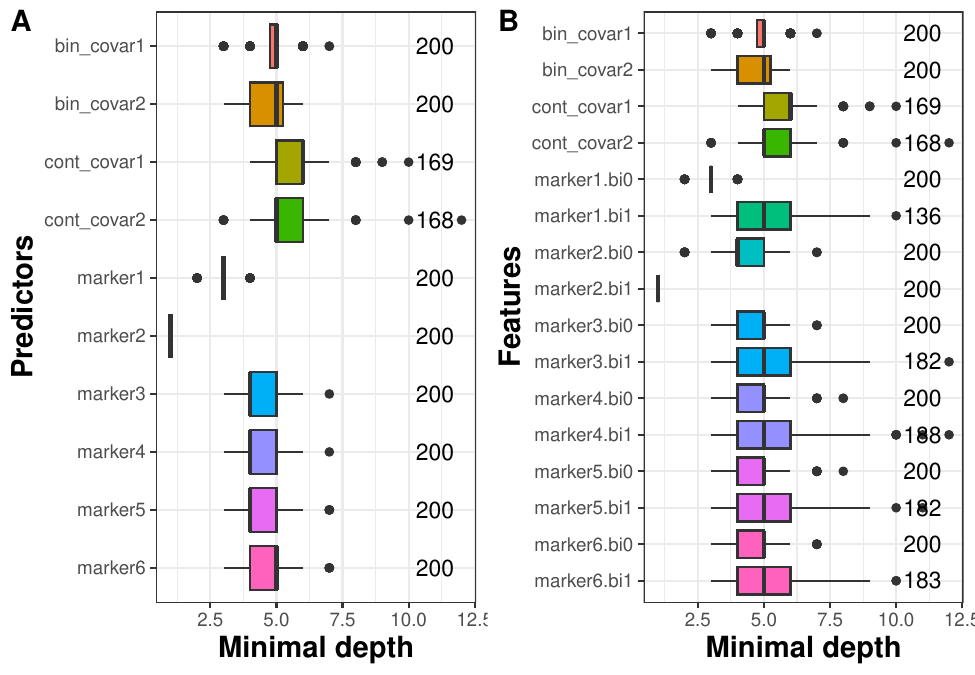} 

}

\caption{Average minimal depth level by predictor (A) and by feature (B).}\label{fig:DynForestRdepthscalar}
\end{figure}

Minimal depth information can be extracted using \texttt{var\_depth()}
function and can be displayed with \texttt{plot()} function. For the
purpose of this illustration, we displayed the minimal depth in figure
\ref{fig:DynForestRdepthscalar} by predictor and by feature.

\begin{Shaded}
\begin{Highlighting}[]
\NormalTok{depth\_dyn }\OtherTok{\textless{}{-}} \FunctionTok{var\_depth}\NormalTok{(}\AttributeTok{DynForest\_obj =}\NormalTok{ res\_dyn)}
\NormalTok{p1 }\OtherTok{\textless{}{-}} \FunctionTok{plot}\NormalTok{(depth\_dyn, }\AttributeTok{plot\_level =} \StringTok{"predictor"}\NormalTok{)}
\end{Highlighting}
\end{Shaded}

\begin{Shaded}
\begin{Highlighting}[]
\NormalTok{p2 }\OtherTok{\textless{}{-}} \FunctionTok{plot}\NormalTok{(depth\_dyn, }\AttributeTok{plot\_level =} \StringTok{"feature"}\NormalTok{)}
\end{Highlighting}
\end{Shaded}

We observe in figure \ref{fig:DynForestRdepthscalar}A that marker2 and
marker1 have the lowest minimal depth, as expected. To go further, we
also looked into the minimal depth computed on features. We perfectly
identified the random slope of marker2 (i.e., marker2.bi1) and the
random intercept of marker1 (i.e., marker1.bi0) as the predictors in
this simulated dataset.

\begin{Shaded}
\begin{Highlighting}[]
\FunctionTok{plot\_grid}\NormalTok{(p1, p2, }\AttributeTok{labels =} \FunctionTok{c}\NormalTok{(}\StringTok{"A"}\NormalTok{, }\StringTok{"B"}\NormalTok{))}
\end{Highlighting}
\end{Shaded}

\hypertarget{discussion}{%
\section{Discussion}\label{discussion}}

The \textbf{DynForest} R package provides an easy-to-use random forests
methodology for predictors that may contain longitudinal variables
possibly measured irregularly with error. Note that the method can also
be used without any longitudinal predictors such as other random forests
packages.

We implemented several statistics to identify the predictive ability of
each variable with the VIMP, gVIMP and average minimal depth. For
survival outcome, compared to \textbf{randomForestSRC} R package, we
considered two different stopping criteria \texttt{nodesize} and
\texttt{minsplit} to favor the deepest forests possible and avoid
suboptimal splits. We designed \textbf{DynForest} to be as user-friendly
as possible. To achieve that, we implemented various functions to
summarize and display the results, and provided a step-by-step analysis
in the three modes; survival, categorical and continuous.

Nevertheless, several improvements could be considered in the future. We
used linear mixed models for longitudinal continuous outcomes but
alternatives strategies could be considered such as PACE algorithm (Yao,
Müller, and Wang 2005) based on functional data analysis. We could also
consider different natures of longitudinal predictors (e.g., binary) for
which generalized linear mixed models could be used. \textbf{DynForest}
currently handles continuous, categorical and survival (with possibly
competing events) outcomes. But other outcomes could be envisaged such
as curves, recurrent events or interval-censored time-to-events. We
leave these perspectives for future releases.

\hypertarget{computational-details}{%
\section*{Computational details}\label{computational-details}}
\addcontentsline{toc}{section}{Computational details}

The results in this paper were obtained using R 4.3.3 with the
\textbf{DynForest} 1.1.3 package on a virtualized Windows Server 2016
Remote Desktop Server with 48GB RAM. R itself and all packages used are
available from the Comprehensive R Archive Network (CRAN) at
\url{https://CRAN.R-project.org/}.

\hypertarget{acknowledgments}{%
\section*{Acknowledgments}\label{acknowledgments}}
\addcontentsline{toc}{section}{Acknowledgments}

We thank Dr.~Louis Capitaine for FrechForest R code used in
\textbf{DynForest}.

This work was funded by the French National Research Agency
(ANR-18-CE36-0004-01 for project DyMES), and the French government in
the framework of the PIA3 (``Investment for the future'') (project
reference 17-EURE-0019) and in the framework of the University of
Bordeaux's IdEx ``Investments for the Future'' program / RRI PHDS.

\hypertarget{references}{%
\section*{References}\label{references}}

\hypertarget{refs}{}
\begin{CSLReferences}{1}{0}
\leavevmode\vadjust pre{\hypertarget{ref-aalen_nonparametric_1976}{}}%
Aalen, Odd. 1976. {``Nonparametric {Inference} in {Connection} with
{Multiple} {Decrement} {Models}.''} \emph{Scandinavian Journal of
Statistics} 3 (1): 15--27. \url{https://www.jstor.org/stable/4615603}.

\leavevmode\vadjust pre{\hypertarget{ref-aalen_empirical_1978}{}}%
Aalen, Odd O., and Søren Johansen. 1978. {``An {Empirical} {Transition}
{Matrix} for {Non}-{Homogeneous} {Markov} {Chains} {Based} on {Censored}
{Observations}.''} \emph{Scandinavian Journal of Statistics} 5 (3):
141--50. \url{https://www.jstor.org/stable/4615704}.

\leavevmode\vadjust pre{\hypertarget{ref-bernard_influence_2009}{}}%
Bernard, Simon, Laurent Heutte, and Sébastien Adam. 2009. {``Influence
of Hyperparameters on Random Forest Accuracy.''} In \emph{International
Workshop on Multiple Classifier Systems}, 171--80. Springer.
https://doi.org/\url{https://doi.org/10.1007/978-3-642-02326-2_18}.

\leavevmode\vadjust pre{\hypertarget{ref-breiman_random_2001}{}}%
Breiman, Leo. 2001. {``Random Forests.''} \emph{Machine Learning} 45
(1): 5--32. \url{https://doi.org/10.1023/A:1010933404324}.

\leavevmode\vadjust pre{\hypertarget{ref-chen_xgboost_2016}{}}%
Chen, Tianqi, and Carlos Guestrin. 2016. {``Xgboost: A Scalable Tree
Boosting System.''} In \emph{Proceedings of the 22nd Acm Sigkdd
International Conference on Knowledge Discovery and Data Mining},
785--94. https://doi.org/\url{https://doi.org/10.1145/2939672.2939785}.

\leavevmode\vadjust pre{\hypertarget{ref-devaux_dynforest_2024}{}}%
Devaux, Anthony. 2024. \emph{{DynForest}: {Random} {Forest} with
{Multivariate} {Longitudinal} {Predictors}}.
\url{https://CRAN.R-project.org/package=DynForest}.

\leavevmode\vadjust pre{\hypertarget{ref-devaux_random_2023}{}}%
Devaux, Anthony, Catherine Helmer, Robin Genuer, and Cécile Proust-Lima.
2023. {``Random Survival Forests with Multivariate Longitudinal
Endogenous Covariates.''} \emph{Statistical Methods in Medical Research}
32 (12): 2331--46. \url{https://doi.org/10.1177/09622802231206477}.

\leavevmode\vadjust pre{\hypertarget{ref-gerds_consistent_2006}{}}%
Gerds, Thomas A., and Martin Schumacher. 2006. {``Consistent
{Estimation} of the {Expected} {Brier} {Score} in {General} {Survival}
{Models} with {Right}-{Censored} {Event} {Times}.''} \emph{Biometrical
Journal} 48 (6): 1029--40. \url{https://doi.org/10.1002/bimj.200610301}.

\leavevmode\vadjust pre{\hypertarget{ref-gray_cmprsk_2020}{}}%
Gray, Bob. 2020. \emph{{cmprsk}: Subdistribution Analysis of Competing
Risks}. \url{https://CRAN.R-project.org/package=cmprsk}.

\leavevmode\vadjust pre{\hypertarget{ref-gray_class_1988}{}}%
Gray, Robert J. 1988. {``A {Class} of {K}-{Sample} {Tests} for
{Comparing} the {Cumulative} {Incidence} of a {Competing} {Risk}.''}
\emph{The Annals of Statistics} 16 (3): 1141--54.
\url{https://www.jstor.org/stable/2241622}.

\leavevmode\vadjust pre{\hypertarget{ref-gregorutti_correlation_2017}{}}%
Gregorutti, Baptiste, Bertrand Michel, and Philippe Saint-Pierre. 2017.
{``Correlation and Variable Importance in Random Forests.''}
\emph{Statistics and Computing} 27 (3): 659--78.
https://doi.org/\url{https://doi.org/10.1007/s11222-016-9646-1}.

\leavevmode\vadjust pre{\hypertarget{ref-ishwaran_random_2014}{}}%
Ishwaran, Hemant, Thomas A. Gerds, Udaya B. Kogalur, Richard D. Moore,
Stephen J. Gange, and Bryan M. Lau. 2014. {``Random Survival Forests for
Competing Risks.''} \emph{Biostatistics} 15 (4): 757--73.
\url{https://doi.org/10.1093/biostatistics/kxu010}.

\leavevmode\vadjust pre{\hypertarget{ref-ishwaran_random_2008}{}}%
Ishwaran, Hemant, Udaya B. Kogalur, Eugene H. Blackstone, and Michael S.
Lauer. 2008. {``Random Survival Forests.''} \emph{The Annals of Applied
Statistics} 2 (3): 841--60. \url{https://doi.org/10.1214/08-AOAS169}.

\leavevmode\vadjust pre{\hypertarget{ref-ishawaran_fast_2022}{}}%
Ishwaran, H., and U. B. Kogalur. 2022. \emph{Fast Unified Random Forests
for Survival, Regression, and Classification (RF-SRC)}.
\url{https://cran.r-project.org/package=randomForestSRC}.

\leavevmode\vadjust pre{\hypertarget{ref-laird_random_effects_1982}{}}%
Laird, Nan M., and James H. Ware. 1982. {``Random-{Effects} {Models} for
{Longitudinal} {Data}.''} \emph{Biometrics} 38 (4): 963--74.
\url{https://doi.org/10.2307/2529876}.

\leavevmode\vadjust pre{\hypertarget{ref-murtaugh_primary_1994}{}}%
Murtaugh, Paul A., E. Rolland Dickson, Gooitzen M. Van Dam, Michael
Malinchoc, Patricia M. Grambsch, Alice L. Langworthy, and Chris H. Gips.
1994. {``Primary Biliary Cirrhosis: {Prediction} of Short-Term Survival
Based on Repeated Patient Visits.''} \emph{Hepatology} 20 (1): 126--34.
\url{https://doi.org/10.1002/hep.1840200120}.

\leavevmode\vadjust pre{\hypertarget{ref-nelson_hazard_1969}{}}%
Nelson, Wayne. 1969. {``Hazard {Plotting} for {Incomplete} {Failure}
{Data}.''} \emph{Journal of Quality Technology} 1 (1): 27--52.
\url{https://doi.org/10.1080/00224065.1969.11980344}.

\leavevmode\vadjust pre{\hypertarget{ref-peto_asymptotically_1972}{}}%
Peto, Richard, and Julian Peto. 1972. {``Asymptotically Efficient Rank
Invariant Test Procedures.''} \emph{Journal of the Royal Statistical
Society: Series A (General)} 135 (2): 185--98.
https://doi.org/\url{https://doi.org/10.2307/2344317}.

\leavevmode\vadjust pre{\hypertarget{ref-proust_lima_estimation_2017}{}}%
Proust-Lima, Cécile, Viviane Philipps, and Benoit Liquet. 2017.
{``Estimation of {Extended} {Mixed} {Models} {Using} {Latent} {Classes}
and {Latent} {Processes}: {The} {R} {Package} {lcmm}.''} \emph{Journal
of Statistical Software} 78 (2): 1--56.
\url{https://doi.org/10.18637/jss.v078.i02}.

\leavevmode\vadjust pre{\hypertarget{ref-sene_individualized_2016}{}}%
Sène, Mbéry, Jeremy MG Taylor, James J Dignam, Hélène Jacqmin-Gadda, and
Cécile Proust-Lima. 2016. {``Individualized Dynamic Prediction of
Prostate Cancer Recurrence with and Without the Initiation of a Second
Treatment: Development and Validation.''} \emph{Statistical Methods in
Medical Research} 25 (6): 2972--91.
https://doi.org/\url{https://doi.org/10.1177/096228021453576}.

\leavevmode\vadjust pre{\hypertarget{ref-shannon_mathematical_1948}{}}%
Shannon, Claude Elwood. 1948. {``A Mathematical Theory of
Communication.''} \emph{The Bell System Technical Journal} 27 (3):
379--423. \url{https://doi.org/10.1002/j.1538-7305.1948.tb01338.x}.

\leavevmode\vadjust pre{\hypertarget{ref-therneau_2022_survival}{}}%
Therneau, Terry M. 2022. \emph{A Package for Survival Analysis in {R}}.
\url{https://CRAN.R-project.org/package=survival}.

\leavevmode\vadjust pre{\hypertarget{ref-wright_ranger_2017}{}}%
Wright, Marvin N., and Andreas Ziegler. 2017. {``Ranger: {A} {Fast}
{Implementation} of {Random} {Forests} for {High} {Dimensional} {Data}
in {C++} and {R}.''} \emph{Journal of Statistical Software} 77 (1).
\url{https://doi.org/10.18637/jss.v077.i01}.

\leavevmode\vadjust pre{\hypertarget{ref-yao_functional_2005}{}}%
Yao, Fang, Hans-Georg Müller, and Jane-Ling Wang. 2005. {``Functional
{Data} {Analysis} for {Sparse} {Longitudinal} {Data}.''} \emph{Journal
of the American Statistical Association} 100 (470): 577--90.
\url{https://doi.org/10.1198/016214504000001745}.

\end{CSLReferences}

\bibliographystyle{unsrt}
\bibliography{RJreferences.bib}

\end{document}